\pdfoutput=1  
\documentclass[letterpaper]{article} %
\usepackage[preprint]{aaai2027}
\usepackage[hyphens]{url}  %
\usepackage{graphicx} %
\usepackage{natbib}  %
\usepackage{caption} %
\usepackage{booktabs}
\usepackage{multirow}
\usepackage{tabularx}
\renewcommand{\topfraction}{0.92}
\renewcommand{\textfraction}{0.06}
\renewcommand{\floatpagefraction}{0.75}
\renewcommand{\dbltopfraction}{0.92}
\renewcommand{\dblfloatpagefraction}{0.7}
\newcolumntype{C}{>{\centering\arraybackslash}X}
\newcolumntype{L}{>{\raggedright\arraybackslash}X}
\newcolumntype{G}{>{\color{ggrey}}c}

\usepackage{amsmath,amssymb}
\usepackage{xcolor}
\usepackage{colortbl}
\usepackage{xspace}

\usepackage{pgfplots}
\pgfplotsset{compat=1.18}
\usetikzlibrary{arrows.meta,positioning,decorations.pathreplacing,calc}
\usepgfplotslibrary{fillbetween}
\usepgfplotslibrary{groupplots}
\definecolor{gblue}{HTML}{4285F4}
\definecolor{gred}{HTML}{EA4335}
\definecolor{ggreen}{HTML}{34A853}
\definecolor{ggrey}{HTML}{9AA0A6}
\definecolor{ggreyd}{HTML}{5F6368}
\definecolor{ggrid}{HTML}{E8EAED}
\definecolor{gyellow}{HTML}{F9AB00}
\definecolor{gpurple}{HTML}{A142F4}
\colorlet{ourrow}{gblue!7}

\usepackage{cleveref}
\crefname{figure}{Fig.}{Figs.}\Crefname{figure}{Fig.}{Figs.}
\crefname{table}{Tab.}{Tabs.}\Crefname{table}{Tab.}{Tabs.}
\crefname{section}{Sec.}{Secs.}\Crefname{section}{Sec.}{Secs.}
\crefname{subsection}{Sec.}{Secs.}\Crefname{subsection}{Sec.}{Secs.}

\newcommand{\method}{\textsc{RTI}\xspace}          %
\newcommand{\Read}{\textsc{Read}\xspace}
\newcommand{\Write}{\textsc{Write}\xspace}
\newcommand{\Rcells}{R}                             %
\newcommand{\Ntok}{N}                               %
\newcommand{\xt}{\mathbf{x}_t}

\newcommand{\papertitle}{Elastic Token Compression for Pixel-Space Diffusion Transformers}
\title{\papertitle}

\author{
    Eduard Zamfir\textsuperscript{\rm 1}\quad
    Christian Reisswig\textsuperscript{\rm 2}\quad
    Zongwei Wu\textsuperscript{\rm 1}\quad
    Yongqin Xian\textsuperscript{\rm 2}\quad
    Radu Timofte\textsuperscript{\rm 1}
}
\affiliations{
    \textsuperscript{\rm 1}University of W\"urzburg \quad
    \textsuperscript{\rm 2}Google\\
}

\ifdefined\appendixonly
  \usepackage{xr}
\fi

\begin{document}

\ifdefined\appendixonly\else
\maketitle

\begin{abstract}
Natural images concentrate their detail in a small fraction of the frame, yet diffusion
models spend a full token on every patch, in every layer and at every timestep.
The waste is largest in pixel-space models, with no autoencoder to absorb low-level
redundancy first. Probing a pretrained pixel text-to-image transformer, we find its
middle-block tokens redundant wherever the image is flat. The redundancy occupies connected,
content-shaped regions, and exploiting it requires tokens with the same geometry. Cutting a Hilbert ordering of the patches provides them. Consecutive positions are
always image neighbours, so any contiguous run is a connected region
whose size and shape follow the content, and grouping in two dimensions becomes a cut in
one. Existing reductions each lose part of this. Similarity merging scatters its groups,
latent bottlenecks discard position, and skipping deletes what it should summarize. We cut where the model's features change most and pool each run into one
region token. Our Region Token Interface (\method{}) adapts a diffusion model to
these tokens, with the region count drawn at random during fine-tuning so one
checkpoint serves every budget. \method{} leads prior reduction methods at matched budgets, matches dense quality at
$2.0\times$ the speed, and stays close at $2.6\times$. The code and models are open-sourced
at \url{https://eduardzamfir.github.io/rti}.
\end{abstract}

\section{Introduction}
\label{sec:intro}

\begin{figure}[!t]
\centering
\input{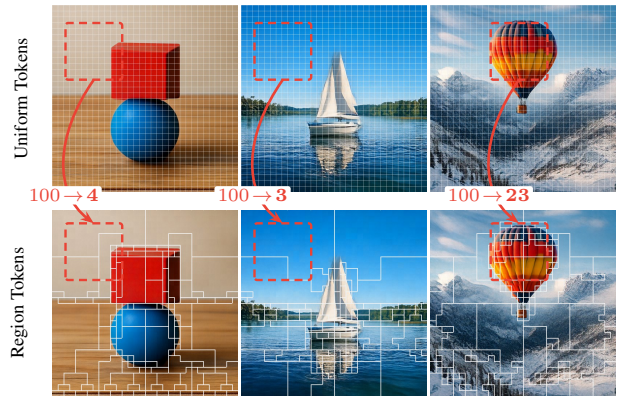}
\caption{\textbf{Content-adaptive tokenization.} \method{} groups the patch grid into
contiguous regions sized by content, so flat areas collapse into a few large region tokens
while detail stays near per-patch. Three samples with the uniform grid overlaid (top) and
ours (bottom); each arrow reports how many region tokens \method{} spends where the grid
spends {\color{gred}${\sim}100$} patches. The overlays compare \emph{tokenizations}, not outputs.}
\label{fig:teaser}
\end{figure}

A natural image is not uniform in space, and the process that generates it is not uniform in
time. Detail gathers in a small part of the frame while clear sky, a flat tabletop or a
stretch of water repeats itself over large areas, and a diffusion
model~\citep{hoDenoisingDiffusionProbabilistic2020,song2021ddim} resolves that detail only
late in sampling, once the layout has settled. Diffusion
transformers~\citep{peebles2023dit,ma2024sit,esser2024sd3} follow neither axis. They run on a
regular grid of patches, one token each, at the same cost in every layer and at every
timestep. \cref{fig:teaser} makes the mismatch concrete. Where a uniform tokenization spends around a
hundred patches on sky or a wall, a handful of larger regions covers the same area.

The waste is most acute in \emph{pixel-space} diffusion
transformers~\citep{liBackBasicsLet2026,minit2i2026}, which denoise the image directly and so
have no autoencoder to absorb the low-level spatial redundancy first. A latent
model~\citep{rombachHighResolutionImageSynthesis2022} pays for that redundancy once, in a
VAE; a pixel-space model pays for it in every block.

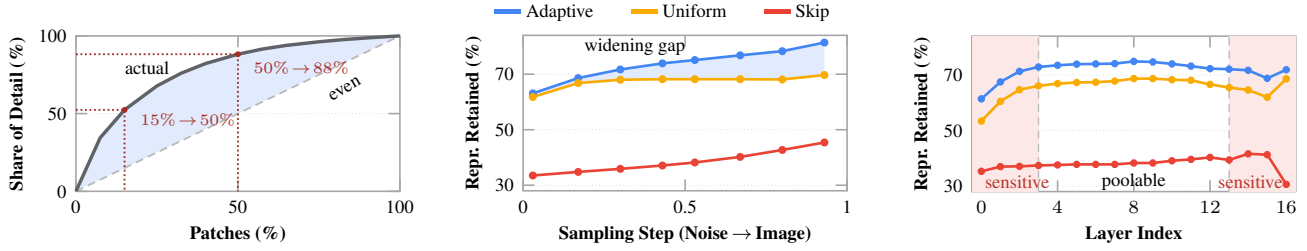
\begin{figure*}[!t]
\centering
\pgfplotsset{poolaxis/.style={
    width=\linewidth, height=0.62\linewidth,
    ylabel style={font=\scriptsize}, xlabel style={font=\scriptsize},
    tick label style={font=\scriptsize}, xlabel shift=-2pt, ylabel shift=-2pt,
    ymajorgrids=true, grid style={line width=0.3pt, draw=gray!30, densely dotted},
    axis line style={line width=0.6pt, draw=gray!70},
    tick style={draw=gray!70, line width=0.6pt}, tick align=inside, minor tick num=0,
    clip=false,
}}
\centering
{\scriptsize\color{black}
  \textcolor{gblue}{\rule[0.35ex]{10pt}{1.6pt}}\;Adaptive\qquad
  \textcolor{gyellow}{\rule[0.35ex]{10pt}{1.6pt}}\;Uniform\qquad
  \textcolor{gred}{\rule[0.35ex]{10pt}{1.6pt}}\;Skip}\\[-1pt]
\begin{minipage}[t]{0.33\textwidth}
\centering
\begin{tikzpicture}
  \begin{axis}[poolaxis, xmin=0, xmax=1, ymin=0, ymax=1,
    xtick={0,0.5,1}, xticklabels={0,50,100}, ytick={0,0.5,1}, yticklabels={0,50,100},
    xlabel={\textbf{Patches (\%)}}, ylabel={\textbf{Share of Detail (\%)}}]
  \addplot[name path=diag, gray!55, densely dashed, line width=0.7pt] coordinates {(0,0) (1,1)};
  \addplot[name path=lor, ggreyd, line width=1.3pt] coordinates {(0,0)(0.075,0.345)(0.15,0.524)(0.25,0.678)(0.325,0.760)(0.40,0.822)(0.50,0.882)(0.575,0.915)(0.65,0.939)(0.75,0.962)(0.825,0.976)(0.90,0.987)(1,1)};
  \addplot[gblue!16, forget plot] fill between[of=lor and diag];
  \node[font=\scriptsize, text=black, rotate=30] at (axis cs:0.83,0.66) {even};
  \node[font=\scriptsize, text=black] at (axis cs:0.22,0.795) {actual};
  \draw[gred!70!black, densely dotted, line width=0.7pt]
    (axis cs:0,0.524) -- (axis cs:0.15,0.524) -- (axis cs:0.15,0);
  \draw[gred!70!black, densely dotted, line width=0.7pt]
    (axis cs:0,0.882) -- (axis cs:0.50,0.882) -- (axis cs:0.50,0);
  \fill[gred!70!black] (axis cs:0.15,0.524) circle (1.1pt);
  \fill[gred!70!black] (axis cs:0.50,0.882) circle (1.1pt);
  \node[font=\tiny, text=gred!70!black, anchor=west] at (axis cs:0.17,0.46) {$15\%\!\to\!50\%$};
  \node[font=\tiny, text=gred!70!black, anchor=west] at (axis cs:0.52,0.80) {$50\%\!\to\!88\%$};
  \end{axis}
\end{tikzpicture}
\par\smallskip
{(a) Detail is spatially concentrated.}
\end{minipage}\hfill
\begin{minipage}[t]{0.33\textwidth}
\centering
\begin{tikzpicture}
  \begin{axis}[poolaxis, ymin=0.28, ymax=0.84, ytick={0.3,0.5,0.7}, yticklabels={30,50,70},
    xmin=0, xmax=1, xtick={0,0.5,1}, xticklabels={0,0.5,1},
    xlabel={\textbf{Sampling Step (Noise\,$\to$\,Image)}}, ylabel={\textbf{Repr.\ Retained (\%)}}]
  \addplot[name path=qt, gblue, line width=1.0pt, mark=*, mark size=1.0pt, mark options={fill=gblue}]
    coordinates {(0.03,0.631)(0.17,0.686)(0.30,0.717)(0.43,0.739)(0.53,0.751)(0.67,0.768)(0.80,0.783)(0.93,0.814)};
  \addplot[name path=un, gyellow, line width=1.0pt, mark=*, mark size=1.0pt, mark options={fill=gyellow}]
    coordinates {(0.03,0.618)(0.17,0.668)(0.30,0.680)(0.43,0.682)(0.53,0.682)(0.67,0.682)(0.80,0.681)(0.93,0.697)};
  \addplot[gblue!16, forget plot] fill between[of=qt and un];
  \addplot[gred, line width=1.0pt, mark=*, mark size=1.0pt, mark options={fill=gred}]
    coordinates {(0.03,0.335)(0.17,0.348)(0.30,0.359)(0.43,0.371)(0.53,0.382)(0.67,0.402)(0.80,0.427)(0.93,0.454)};
  \node[font=\scriptsize, text=black, anchor=west] at (axis cs:0.16,0.80) {widening gap};
  \end{axis}
\end{tikzpicture}
\par\smallskip
{(b) More redundant as detail forms.}
\end{minipage}\hfill
\begin{minipage}[t]{0.33\textwidth}
\centering
\begin{tikzpicture}
  \begin{axis}[poolaxis, ymin=0.28, ymax=0.84, ytick={0.3,0.5,0.7}, yticklabels={30,50,70},
    xmin=-0.5, xmax=16.5, xtick={0,4,8,12,16},
    xlabel={\textbf{Layer Index}}, ylabel={\textbf{Repr.\ Retained (\%)}}]
  \addplot[draw=none, fill=gred!12, forget plot] coordinates {(-0.5,0.28) (-0.5,0.84) (3,0.84) (3,0.28)} \closedcycle;
  \addplot[draw=none, fill=gred!12, forget plot] coordinates {(13,0.28) (13,0.84) (16.5,0.84) (16.5,0.28)} \closedcycle;
  \draw[gray!55, densely dashed, line width=0.5pt] (axis cs:3,0.28) -- (axis cs:3,0.84);
  \draw[gray!55, densely dashed, line width=0.5pt] (axis cs:13,0.28) -- (axis cs:13,0.84);
  \node[font=\scriptsize, text=gred!80!black, anchor=west] at (axis cs:-0.3,0.315) {sensitive};
  \node[font=\scriptsize, text=gred!80!black, anchor=east] at (axis cs:16.3,0.315) {sensitive};
  \node[font=\scriptsize, text=black] at (axis cs:8,0.315) {poolable};
  \addplot[gblue, line width=1.0pt, mark=*, mark size=0.9pt, mark options={fill=gblue}]
    coordinates {(0,0.614)(1,0.675)(2,0.713)(3,0.729)(4,0.735)(5,0.739)(6,0.740)(7,0.741)(8,0.749)(9,0.747)(10,0.740)(11,0.732)(12,0.723)(13,0.721)(14,0.717)(15,0.688)(16,0.719)};
  \addplot[gyellow, line width=1.0pt, mark=*, mark size=0.9pt, mark options={fill=gyellow}]
    coordinates {(0,0.534)(1,0.605)(2,0.647)(3,0.661)(4,0.669)(5,0.673)(6,0.674)(7,0.678)(8,0.687)(9,0.687)(10,0.683)(11,0.681)(12,0.666)(13,0.655)(14,0.646)(15,0.620)(16,0.686)};
  \addplot[gred, line width=1.0pt, mark=*, mark size=0.9pt, mark options={fill=gred}]
    coordinates {(0,0.353)(1,0.370)(2,0.371)(3,0.374)(4,0.376)(5,0.378)(6,0.378)(7,0.378)(8,0.383)(9,0.383)(10,0.391)(11,0.396)(12,0.403)(13,0.394)(14,0.416)(15,0.413)(16,0.306)};
  \end{axis}
\end{tikzpicture}
\par\smallskip
{(c) Most poolable in the middle blocks.}
\end{minipage}
\caption{\textbf{Token redundancy in a pretrained pixel diffusion model.} We probe a frozen
MiniT2I-B/16 along three axes. \textbf{(a)}~In \emph{space}, detail is concentrated: the
most-detailed $15\%$ of patches hold half of it, the top half $88\%$. \textbf{(b)}~In
\emph{time}, pooling the $\Ntok{=}1024$ tokens to $\Rcells{=}256$ regions recovers a growing
share of a middle block's representation as the image forms, with a content-adaptive
partition ahead of a uniform grid throughout. \textbf{(c)}~In \emph{depth}, the redundancy
sits in the middle blocks and thins toward the first and last (shaded); skipping rather than
pooling (red) recovers far less.}
\label{fig:poolability}
\end{figure*}

We measure that redundancy in a pretrained pixel generator, along the two axes above and a
third that the model adds. In \emph{space}, \cref{fig:poolability}a confirms the
concentration in the generated images themselves. In \emph{time}, \cref{fig:poolability}b pools the
$\Ntok$ patch tokens into a few hundred regions and keeps a growing share of a block's
representation as the image forms. In \emph{depth}, \cref{fig:poolability}c finds that share
highest in the middle blocks and falling toward the first and last, which stay sensitive to
every patch.

Measured this way, the redundancy has a shape, and the shape dictates the reduction.
Redundant patches lie next to one another in flat stretches of the frame, so groups must be
\emph{contiguous}. Because detail concentrates in few patches, group size must follow the
\emph{content}, large over a flat wall and small over foliage. A group covers a definite
place that its token has to stand for, and each must therefore hold a \emph{position}. And
in \cref{fig:poolability}c dropping loses what pooling keeps, which is why the redundant
content must be \emph{summarized} rather than deleted, coarse rather than absent. 
Existing reductions each forfeit at least one. Similarity
merging~\citep{bolya2023tome,bolya2023tomesd,wang2024atedm} groups by appearance, so a group
can gather patches from opposite corners of the frame and its token stands nowhere in
particular, losing contiguity and position at once; latent
bottlenecks~\citep{jaegle2022perceiverio,jabri2023rin,hajiali2026elit} give up grouping and
position together; conditional
computation~\citep{raposo2024mod,jainMixtureNestedExperts2024} deletes what it should
summarize.

In this work, we obtain all of these properties from a single ordering of the patches. A
Hilbert curve~\citep{hilbert1891} visits every patch exactly once, and consecutive positions
in the order it induces are always image neighbours, so any contiguous run of that order is a
connected region, of any length and not necessarily a square, with a centre the pretrained
attention reads as an ordinary position.
Grouping in two dimensions therefore becomes cutting in one.
We cut where the model's features change most, long runs over flat areas, single patches
over detail, and recut each step, so the partition tracks the image as it forms.

\begin{figure*}[!t]
\centering
\input{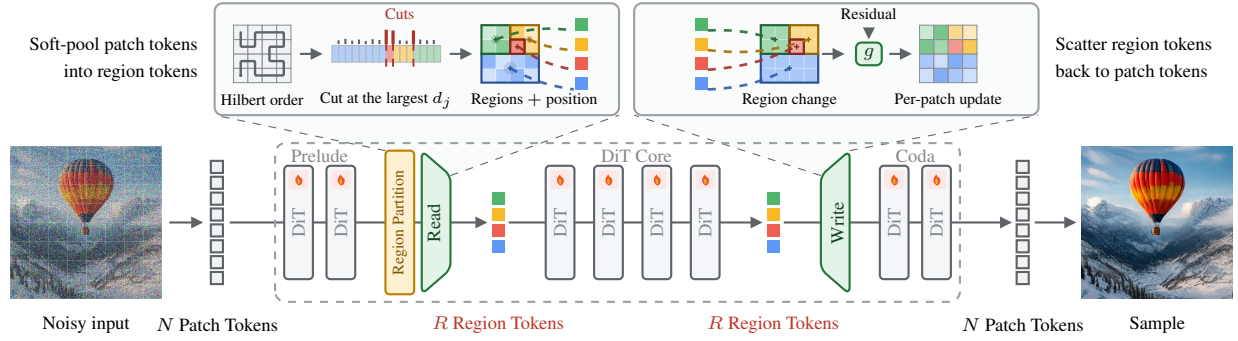}
\caption{\textbf{\method{}: an elastic region-token interface.} The first and last blocks
keep all $\Ntok$ patch tokens; the core between them runs on $\Rcells \ll \Ntok$ region
tokens. \emph{Left:} cuts at the $\Rcells{-}1$ largest feature gaps $d_j$ (red) along the
Hilbert order give the regions, and \Read{} pools each into one token with its size and
position. \emph{Right:} \Write{} returns each region token's change to its
patches, a learned map $g$ specializing the update with the patch's own pre-pool features.
Only the interface and adapters train (grey: LoRA; green: \Read{}/\Write{}); the budget is
drawn at random, so one checkpoint serves every budget.}
\label{fig:method}
\end{figure*}

We integrate this partition into a pretrained diffusion transformer with \method{}, our
\textit{Region Token Interface}. Guided by \cref{fig:poolability}c, the first and last blocks keep
every patch, while the middle blocks operate on \emph{region tokens} alone. A learned pooling forms
each region token on the way in, and a learned scattering spreads its update back over the
patches on the way out. The backbone stays
frozen apart from low-rank adapters, and the number of regions is drawn at random during
training, so one checkpoint runs at any budget.

In summary, our main contributions are:
\begin{itemize}
\setlength{\itemsep}{1pt}\setlength{\parskip}{0pt}
\item We measure the redundancy in a pretrained pixel diffusion transformer and the shape
it takes.
\item We introduce \method{}, an elastic Hilbert-ordered region interface that adapts a frozen backbone.
\item \method{} matches the dense baseline at twice the throughput, ahead of prior reduction methods.
\end{itemize}

\section{Related Work}
\label{sec:related}

\paragraph{Cheaper diffusion transformers.} Most work on making diffusion
transformers~\citep{peebles2023dit,ma2024sit,esser2024sd3} cheaper reduces the number of
\emph{sampling steps}, by distillation, consistency
models~\citep{salimans2022progressive,song2023consistency,luo2023lcm} or caching across
steps~\citep{ma2024deepcache}. It leaves the per-step token cost untouched, which is what we reduce.

\paragraph{Token reduction and adaptive tokenization.} Token Merging~\citep{bolya2023tome},
adapted to diffusion~\citep{bolya2023tomesd}, groups similar tokens within a forward
pass; later variants prune by importance and vary the ratio across layers and
timesteps~\citep{wang2024atedm,smith2024todo,wu2025importancetome,you2025diffcr}. All
group by feature similarity or importance rather than by where a token sits. A separate line
makes the tokenizer itself content-adaptive in \emph{input space}, by quadtree or
native-resolution tokenization~\citep{ronen2023quadformer,dehghani2023navit,qdm2025}, but
trains it from scratch, which rules it out for a released checkpoint. Closest to us,
DC-DiT~\citep{dynchunk2026} can likewise be upcycled from a pretrained transformer, but its
chunking is router-learned rather than parameter-free, targets an average ratio rather than
an exact budget, and works in a VAE latent space, which has already absorbed much of the
redundancy we measure in pixels. Neither content-adaptive tokenization nor the Hilbert order is new; point-cloud
transformers use the curve to order the \emph{sequence} that attention runs
over~\citep{wang2023pointtransformerv3}; here it defines the \emph{groups} themselves.

\paragraph{Latent bottlenecks and skipping.} Our hourglass shape is that of hierarchical
transformers~\citep{nawrot2022hourglass,geiping2025recurrentdepth}, which pool by fixed
stride; we replace the pooling with a content-adaptive partition on a frozen checkpoint.
Cross-attention bottlenecks instead compress into learned latents and read them back:
Perceiver-IO~\citep{jaegle2022perceiverio}, Set-Transformer inducing
points~\citep{lee2019settransformer}, soft mixtures of
experts~\citep{puigcerver2024softmoe}, and, closest, ELIT's elastic latent
interface~\citep{hajiali2026elit}. Our \Read{}/\Write{} belongs to that family but anchors
its latents spatially: a read stays inside a region, costing $\mathcal{O}(\Ntok)$ rather than
the $\mathcal{O}(\Ntok\Rcells)$ of global queries, and every region token carries a position
the frozen rotary attention reads. Mixture-of-Depths and per-token
routing~\citep{raposo2024mod,jainMixtureNestedExperts2024} take the opposite route, skipping
tokens rather than summarizing them.

\section{Method}
\label{sec:method}

\subsection{Preliminaries}
\label{sec:prelim}

\paragraph{Base model.} We build on MiniT2I-B/16~\citep{minit2i2026}, a pixel-space
(JiT-family) text-to-image diffusion transformer whose double-stream trunk of $17$ blocks
carries image tokens alongside frozen FLAN-T5 text tokens. At $512^2$ resolution and patch
size $16$, an image becomes a $32{\times}32$ grid of $\Ntok{=}1024$ patch tokens; this is the
sequence \method{} shortens. The model is trained as a rectified flow, with a state $\xt$
that interpolates between noise and the clean image, and is supervised to predict the clean
image $\hat x$ rather than a noise residual~\citep{liBackBasicsLet2026}.

\paragraph{Hilbert order.} A Hilbert curve~\citep{hilbert1891} visits every cell of a
$2^k{\times}2^k$ grid exactly once, and consecutive positions along it are always neighbours
in the grid. Any contiguous run of positions therefore occupies a connected region of the
image. Stronger still, every dyadic (quadtree) cell is exactly one such run. Cutting the
sequence can therefore express any quadtree tiling, and tilings no quadtree can reach. We
write $\pi$ for the order the curve induces on the $32{\times}32$ patch grid, with $\pi(j)$
the patch at position $j$.

\subsection{Compressing Only the Core}
\label{sec:core}

\method{} changes only how image tokens are formed and processed inside the trunk. Following
the prelude/core/coda structure of recurrent-depth
transformers~\citep{geiping2025recurrentdepth}, we designate a contiguous span of middle
blocks $[s_0,s_1]$ as the \emph{core}. As \cref{fig:method} shows, a learned \Read{} before
$s_0$ pools the $\Ntok$ patch tokens into $\Rcells \ll \Ntok$ \emph{region tokens}, every
block of the core runs on that one short sequence, and a learned \Write{} after $s_1$
scatters the result back to the $\Ntok$ patches; the blocks outside the core stay at full
resolution. We fit the trunk to the shorter sequence with LoRA~\citep{hu2022lora} on its
attention and MLP projections; the text stream always runs at full length.

The ends resist compression for opposite reasons. The prelude is still turning raw patches
into features, and pooling there would discard pixels before the model has made anything of
them; the coda turns features back into per-pixel output, which a shared region update would
blur. Between them the features are most redundant. The share of a block's representation that
survives the pooling rises from $61\%$ at the first block to ${\sim}75\%$ through the middle
span and falls again toward the last (\cref{fig:poolability}c).

\begin{table*}[!t]
\centering
\footnotesize
\setlength{\tabcolsep}{2.0pt}
\begin{tabularx}{\textwidth}{@{}L ccc *{6}{G} cccc@{}}
\toprule
 & & & & \multicolumn{7}{c}{\emph{GenEval accuracy (\%)\,$\uparrow$}} & \multicolumn{3}{c}{\emph{PartiPrompts\,$\uparrow$}} \\
\cmidrule(lr){5-11}\cmidrule(lr){12-14}
Model & $\Rcells$ & img/s\,$\uparrow$ & Speed\,$\uparrow$ & sng & two & cnt & col & pos & att & \textbf{Ovr.} & CLIP & Pick & IR \\
\midrule
\emph{MiniT2I-B/16}~\citep{minit2i2026} & $1024$ & $0.40$ & $1.00\times$ & 99.4 & 96.0 & 77.8 & 92.8 & 79.5 & 78.0 & $87.2$ & $28.23$ & $22.45$ & $1.18$ \\
\quad Token Skip$_{\text{\tiny\citep{raposo2024mod}}}$ & $576^{\ast}$ & $0.48$ & $1.19\times$ & 98.1 & 91.2 & 65.3 & 90.2 & 77.3 & 71.5 & $82.3_{\text{\tiny$\pm1.7$}}$ & $28.08_{\text{\tiny\color{ggrey}$\pm.10$}}$ & $21.93_{\text{\tiny\color{ggrey}$\pm.03$}}$ & $1.06_{\text{\tiny\color{ggrey}$\pm.02$}}$ \\
\addlinespace[3pt]
\quad Feat Sim$_{\text{\tiny\citep{bolya2023tome}}}$ & $512$ & $0.74$ & $1.84\times$ & 98.4 & 91.9 & 72.5 & 90.2 & 79.8 & 75.5 & $84.7_{\text{\tiny\color{ggrey}$\pm1.6$}}$ & $28.04_{\text{\tiny\color{ggrey}$\pm.09$}}$ & $22.04_{\text{\tiny\color{ggrey}$\pm.02$}}$ & $1.07_{\text{\tiny\color{ggrey}$\pm.02$}}$ \\
\rowcolor{ourrow}\quad \method{} (ours) & $512$ & $0.74$ & $1.84\times$ & 99.7 & 96.0 & 77.8 & 91.2 & 81.5 & 80.8 & $\mathbf{87.8}_{\text{\tiny\color{ggrey}$\pm1.5$}}$ & $28.09_{\text{\tiny\color{ggrey}$\pm.08$}}$ & $22.30_{\text{\tiny\color{ggrey}$\pm.02$}}$ & $1.11_{\text{\tiny\color{ggrey}$\pm.01$}}$ \\
\addlinespace[3pt]
\quad Feat Sim$_{\text{\tiny\citep{bolya2023tome}}}$ & $256$ & $0.86$ & $2.15\times$ & 99.1 & 86.6 & 60.9 & 89.4 & 76.0 & 71.0 & $80.5_{\text{\tiny$\pm1.9$}}$ & $27.73_{\text{\tiny\color{ggrey}$\pm.11$}}$ & $21.57_{\text{\tiny\color{ggrey}$\pm.03$}}$ & $0.93_{\text{\tiny\color{ggrey}$\pm.02$}}$ \\
\quad Latent Array$_{\text{\tiny\citep{jaegle2022perceiverio}}}$ & $256$ & $0.86$ & $2.15\times$ & 86.6 & 39.1 & 28.1 & 72.1 & 27.8 & 20.8 & $45.7_{\text{\tiny\color{ggrey}$\pm3.1$}}$ & $23.30_{\text{\tiny\color{ggrey}$\pm.23$}}$ & $19.78_{\text{\tiny\color{ggrey}$\pm.07$}}$ & $-0.65_{\text{\tiny\color{ggrey}$\pm.06$}}$ \\
\rowcolor{ourrow}\quad \method{} (ours) & $256$ & $0.86$ & $\mathbf{2.15\times}$ & 99.1 & 92.7 & 70.9 & 89.4 & 80.8 & 75.2 & $\mathbf{84.7}_{\text{\tiny\color{ggrey}$\pm1.7$}}$ & $28.04_{\text{\tiny\color{ggrey}$\pm.10$}}$ & $\mathbf{22.06}_{\text{\tiny\color{ggrey}$\pm.03$}}$ & $\mathbf{1.06}_{\text{\tiny\color{ggrey}$\pm.02$}}$ \\
\midrule
\emph{MiniT2I-L/16}~\citep{minit2i2026} & $1024$ & $0.14$ & $1.00\times$ & 99.7 & 94.2 & 76.2 & 93.4 & 85.5 & 79.8 & $88.1$ & $28.50$ & $22.75$ & $1.23$ \\
\addlinespace[3pt]
\quad Feat Sim$_{\text{\tiny\citep{bolya2023tome}}}$ & $512$ & $0.29$ & $2.11\times$ & 97.5 & 91.9 & 67.5 & 89.1 & 82.5 & 70.2 & $83.1_{\text{\tiny$\pm1.9$}}$ & $\mathbf{28.80}_{\text{\tiny$\pm.09$}}$ & $22.34_{\text{\tiny$\pm.02$}}$ & $1.16_{\text{\tiny$\pm.01$}}$ \\
\rowcolor{ourrow}\quad \method{} (ours) & $512$ & $0.29$ & $2.11\times$ & 99.4 & 95.5 & 77.2 & 90.7 & 85.5 & 76.8 & $\mathbf{87.5}_{\text{\tiny\color{ggrey}$\pm1.3$}}$ & $28.58_{\text{\tiny\color{ggrey}$\pm.12$}}$ & $\mathbf{22.63}_{\text{\tiny\color{ggrey}$\pm.04$}}$ & $\mathbf{1.20}_{\text{\tiny\color{ggrey}$\pm.03$}}$ \\
\addlinespace[3pt]
\quad Feat Sim$_{\text{\tiny\citep{bolya2023tome}}}$ & $256$ & $0.36$ & $2.59\times$ & 96.2 & 85.9 & 55.9 & 82.7 & 74.2 & 63.7 & $76.5_{\text{\tiny$\pm2.3$}}$ & $28.31_{\text{\tiny$\pm.13$}}$ & $21.76_{\text{\tiny$\pm.03$}}$ & $1.01_{\text{\tiny$\pm.03$}}$ \\
\rowcolor{ourrow}\quad \method{} (ours) & $256$ & $0.36$ & $\mathbf{2.59\times}$ & 99.1 & 93.2 & 71.6 & 92.6 & 85.5 & 76.0 & $86.3_{\text{\tiny\color{ggrey}$\pm1.6$}}$ & $28.52_{\text{\tiny\color{ggrey}$\pm.13$}}$ & $22.37_{\text{\tiny\color{ggrey}$\pm.03$}}$ & $1.17_{\text{\tiny\color{ggrey}$\pm.03$}}$ \\
\bottomrule
\end{tabularx}
\caption{\textbf{Text-to-image generation at $512^2$.} Within each block every method
adapts the same frozen backbone under one protocol and differs only in how it reduces
tokens. \method{} leads every reduction at every matched budget on GenEval; on B/16 at
$\Rcells{=}512$ it clears the dense backbone on GenEval and throughput at once, and L/16
holds dense quality on half its tokens and stays within $1.8$ GenEval on a quarter, at
$2.59\times$. Speedup is against the block's backbone row; $\pm$
subscripts give the $95\%$ spread of a paired difference against that row, which therefore
carries none. $^{\ast}$Not a budget: the mean over Token Skip's dense and sparse core
blocks, above any $\Rcells$ here, hence the low throughput.}
\label{tab:main}
\end{table*}

\begin{figure*}[!t]
\centering
\pgfplotsset{
  oursB/.style={gblue, line width=1.3pt, mark=*, mark size=1.4pt},
  oursL/.style={gblue, opacity=0.45, line width=1.3pt, mark=square*, mark size=1.5pt},
  fsB/.style={ggreen, line width=1.1pt, mark=diamond*, mark size=2.1pt},
  fsL/.style={ggreen, opacity=0.45, line width=1.1pt, mark=diamond, mark size=2.4pt},
  denseB/.style={gray!75, densely dashed, line width=0.7pt},
  denseL/.style={gray!75, opacity=0.45, densely dashed, line width=0.7pt},
  starB/.style={black, only marks, mark=star, mark size=3.0pt, line width=0.9pt},
  starL/.style={black, opacity=0.45, only marks, mark=star, mark size=3.0pt, line width=0.9pt},
}
\providecommand{\betterarrow}[4]{%
  \draw[-{Stealth[length=3pt]}, gray!45, line width=0.5pt]
    (axis cs:#1,#2) -- (axis cs:#3,#4)
    node[pos=1, anchor=west, font=\scriptsize\itshape, gray!50, inner sep=1.5pt] {better};}
\centering
{\scriptsize\color{black}
  \textcolor{gblue}{\rule[0.35ex]{10pt}{1.6pt}}\;\method{} B/16\quad
  \textcolor{gblue!45}{\rule[0.35ex]{10pt}{1.6pt}}\;\method{} L/16\quad
  \textcolor{ggreen}{\rule[0.35ex]{10pt}{1.0pt}}\;Feat Sim B/16\quad
  \textcolor{ggreen!45}{\rule[0.35ex]{10pt}{1.0pt}}\;Feat Sim L/16\quad
  \textcolor{gred}{$\blacktriangle$}\;Token Skip\quad
  $\star$\;Dense}\\[2pt]
\begin{tikzpicture}[baseline]
\begin{groupplot}[
  group style={group size=4 by 1, horizontal sep=0.36cm},
  scale only axis=true, width=3.86cm, height=2.05cm,
  ymin=0.97, ymax=3.08, ytick={1.0,1.5,2.0,2.5,3.0}, yticklabels={},
  ymajorgrids=true, grid style={line width=0.3pt, draw=gray!30, densely dotted},
  axis line style={line width=0.6pt, draw=gray!70},
  tick style={draw=gray!70, line width=0.6pt}, tick align=inside, minor tick num=0,
  xtick pos=left, ytick pos=left, xlabel near ticks, ylabel near ticks,
  enlargelimits=false, clip=false,
  tick label style={font=\scriptsize}, label style={font=\scriptsize},
  title style={font=\footnotesize, anchor=north, at={(0.5,-0.50)}},
  ylabel style={anchor=center, at={(-0.22,0.5)}},
]
\nextgroupplot[xmin=56.0, xmax=90.0, xtick={60,70,80},
  yticklabels={1,1.5,2,2.5,3}, ylabel={\textbf{Speedup ($\times$)}},
  xlabel={\textbf{GenEval Acc.\ (\%)}}, title={(a) Compositional accuracy.}]
  \addplot[denseB] coordinates {(87.24,0.97) (87.24,3.08)};
  \addplot[denseL] coordinates {(88.12,0.97) (88.12,3.08)};
  \addplot[oursB] coordinates {
    (71.8,2.35) (79.0,2.30) (81.3,2.28) (84.6,2.21) (84.7,2.15) (86.5,2.00) (87.9,1.84)};
  \addplot[oursL] coordinates {(69.1,3.01) (80.3,2.86) (86.3,2.59) (87.5,2.11)};
  \addplot[fsB] coordinates {(57.9,2.35) (69.8,2.28) (80.5,2.15) (84.7,1.84)};
  \addplot[fsL] coordinates {(76.5,2.59) (83.1,2.11)};   %
  \addplot[gred, only marks, mark=triangle*, mark size=2.7pt] coordinates {(82.4,1.19)};
  \addplot[starB] coordinates {(87.24,1.00)};
  \addplot[starL] coordinates {(88.12,1.00)};
  \betterarrow{57.0}{1.06}{63.0}{1.42}
\nextgroupplot[xmin=25.5, xmax=28.95, xtick={26,27,28}, ylabel=\empty,
  xlabel={\textbf{CLIPScore}}, title={(b) Image--text alignment.}]
  \addplot[denseB] coordinates {(28.23,0.97) (28.23,3.08)};
  \addplot[denseL] coordinates {(28.50,0.97) (28.50,3.08)};
  \addplot[oursB] coordinates {
    (27.01,2.35) (27.68,2.28) (27.93,2.21) (28.04,2.15) (28.03,2.04) (28.04,2.00)
    (28.09,1.84) (28.10,1.79) (28.09,1.61) (28.09,1.44)};
  \addplot[oursL] coordinates {(27.57,3.01) (28.26,2.86) (28.52,2.59) (28.58,2.11)};
  \addplot[fsB] coordinates {(25.70,2.35) (27.73,2.15) (28.04,1.84)};
  \addplot[fsL] coordinates {(28.31,2.59) (28.80,2.11)};   %
  \addplot[gred, only marks, mark=triangle*, mark size=2.7pt] coordinates {(28.08,1.19)};
  \addplot[starB] coordinates {(28.23,1.00)};
  \addplot[starL] coordinates {(28.50,1.00)};
  \betterarrow{25.62}{1.06}{26.20}{1.42}
\nextgroupplot[xmin=20.25, xmax=22.9, xtick={20.5,21.5,22.5}, ylabel=\empty,
  xlabel={\textbf{PickScore}}, title={(c) Human preference.}]
  \addplot[denseB] coordinates {(22.45,0.97) (22.45,3.08)};
  \addplot[denseL] coordinates {(22.75,0.97) (22.75,3.08)};
  \addplot[oursB] coordinates {
    (21.20,2.35) (21.68,2.28) (21.92,2.21) (22.06,2.15) (22.18,2.04) (22.22,2.00)
    (22.30,1.84) (22.32,1.79) (22.35,1.61) (22.36,1.44)};
  \addplot[oursL] coordinates {(21.34,3.01) (21.92,2.86) (22.37,2.59) (22.63,2.11)};
  \addplot[fsB] coordinates {(20.39,2.35) (21.57,2.15) (22.04,1.84)};
  \addplot[fsL] coordinates {(21.76,2.59) (22.34,2.11)};   %
  \addplot[gred, only marks, mark=triangle*, mark size=2.7pt] coordinates {(21.93,1.19)};
  \addplot[starB] coordinates {(22.45,1.00)};
  \addplot[starL] coordinates {(22.75,1.00)};
  \betterarrow{20.33}{1.06}{20.81}{1.42}
\nextgroupplot[xmin=0.12, xmax=1.29, xtick={0.2,0.6,1.0}, ylabel=\empty,
  xlabel={\textbf{ImageReward}}, title={(d) Reward-model preference.}]
  \addplot[denseB] coordinates {(1.181,0.97) (1.181,3.08)};
  \addplot[denseL] coordinates {(1.232,0.97) (1.232,3.08)};
  \addplot[oursB] coordinates {
    (0.713,2.35) (0.941,2.28) (1.031,2.21) (1.062,2.15) (1.085,2.04) (1.097,2.00)
    (1.108,1.84) (1.112,1.79) (1.120,1.61) (1.116,1.44)};
  \addplot[oursL] coordinates {(0.804,3.01) (1.050,2.86) (1.167,2.59) (1.200,2.11)};
  \addplot[fsB] coordinates {(0.175,2.35) (0.933,2.15) (1.075,1.84)};
  \addplot[fsL] coordinates {(1.01,2.59) (1.16,2.11)};   %
  \addplot[gred, only marks, mark=triangle*, mark size=2.7pt] coordinates {(1.064,1.19)};
  \addplot[starB] coordinates {(1.181,1.00)};
  \addplot[starL] coordinates {(1.232,1.00)};
  \betterarrow{0.155}{1.06}{0.370}{1.42}
\end{groupplot}
\end{tikzpicture}
\caption{\textbf{Quality across the budget range.} One elastic \method{} checkpoint per
backbone, scored at every budget: colour is the method, opacity the backbone (full B/16,
faded L/16), and the dashed lines and stars mark each backbone's own dense model.
\textbf{(a)}~GenEval; \textbf{(b--d)}~the three PartiPrompts scores. Read across the two
backbones, the larger one gives up more of its tokens for less: at $2.11\times$ L/16 is
within $0.6$ GenEval of its dense model, where B/16 has lost $2.5$ by $2.15\times$.}
\label{fig:frontier}
\end{figure*}
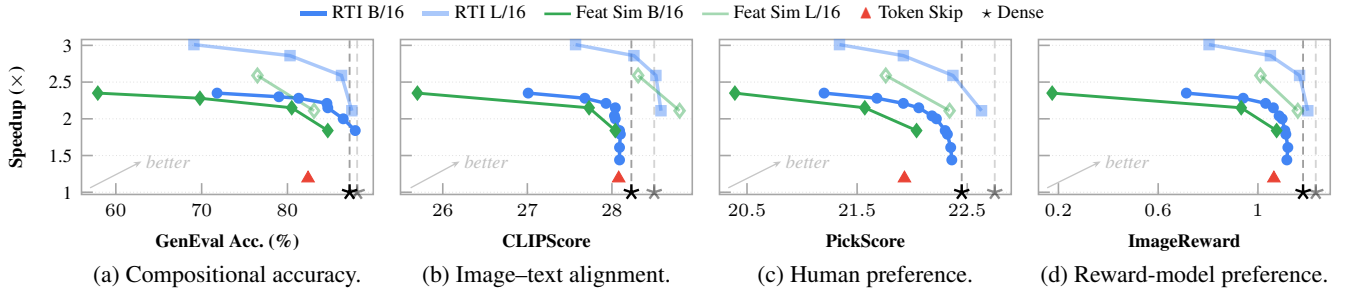

\subsection{Content-Adaptive Region Partition}
\label{sec:partition}

We want $\Rcells$ regions that are connected, coarse over flat areas and fine over detailed
ones, and cheap enough to rebuild at every sampling step. A quadtree delivers the adaptive
sizing but ties every region to a dyadic square and needs a tree built per image per step.
Cutting the Hilbert order keeps the sizing and sheds both drawbacks: the tiling comes from
one pass over a sequence, as \cref{fig:method} illustrates.

Let $\mathbf{h}^{(s_0)}_j \in \mathbb{R}^{d}$ denote the features entering the core. We
score the boundary between neighbouring patches on the curve by their feature distance,
\begin{equation}
d_j = \big\| \mathbf{h}^{(s_0)}_{\pi(j+1)} - \mathbf{h}^{(s_0)}_{\pi(j)} \big\|_2 ,
\qquad j = 1,\dots,\Ntok{-}1 ,
\label{eq:feat_gap}
\end{equation}
cut the sequence at the $\Rcells{-}1$ largest of these,
\begin{equation}
\mathcal{C} = \big\{\, j : d_j \text{ is among the } \Rcells{-}1 \text{ largest} \,\big\} ,
\label{eq:feat_cuts}
\end{equation}
and take the $\Rcells$ runs between consecutive cuts as the regions. Because the cuts land on the
largest feature jumps, homogeneous runs stay merged while detailed areas split down to
single patches.

The rule is parameter-free and costs $\mathcal{O}(\Ntok)$: one
pass for the gaps, a linear selection for the cuts. Since the backbone predicts the clean
image rather than a noise residual, its features track image structure, and a large gap marks
a content boundary. The cut is taken from the current state $\xt$, and in
\cref{fig:routing}a fine regions gather on objects as they resolve, an alignment that
strengthens over sampling (\cref{fig:routing}b). The budget $\Rcells$ is the number of
regions, drawn at random during training from the budget set of \cref{sec:exp-setup}, so
that one model serves any budget.

\begin{figure*}[t]
\centering
\input{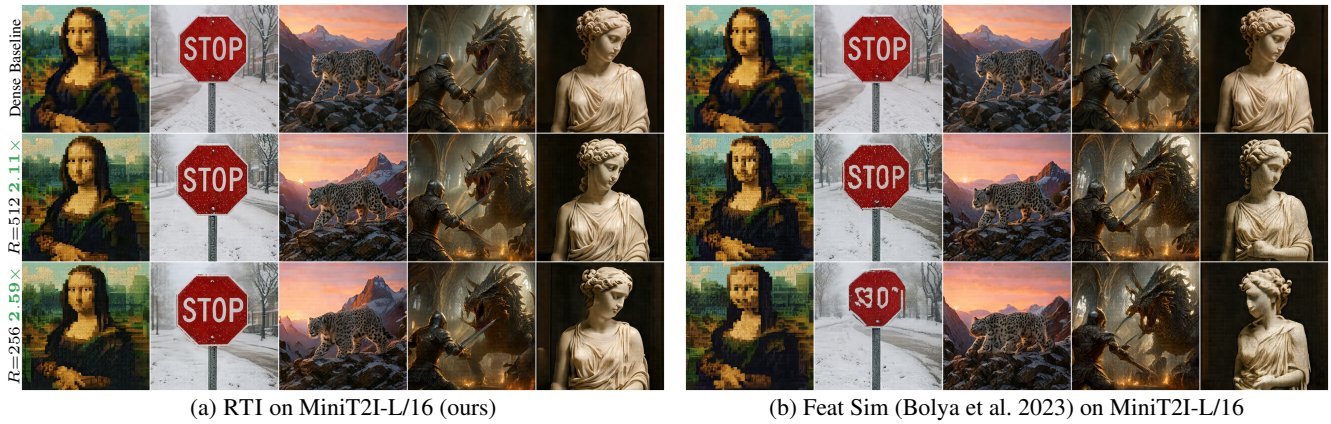}
\caption{\textbf{\method{} against feature-similarity merging at $512^2$.} Five prompts,
one seed. The top row of each panel is the frozen MiniT2I-L/16; below it each method runs at
$\Rcells{=}512$ and $\Rcells{=}256$ at matched throughput, so the panels differ only in how
the tokens are grouped. \method{} holds the dense image at both budgets; Feat Sim comes
apart at $\Rcells{=}256$, where the lettering on the sign stops being letters and the statue
loses its face. \textit{Please refer to the appendix for more samples}.}
\label{fig:gallery}
\end{figure*}

\subsection{The Read / Write Interface}
\label{sec:interface}

Write $c$ for a region, $|c|$ for its number of patches and $c(j)$ for the region containing
patch $j$. \Read{} turns each region into one token and \Write{} turns that token's update
back into per-patch updates. Both act only \emph{within} a region, costing
$\mathcal{O}(\Ntok)$ regardless of $\Rcells$, and between them they hold all the interface's
learned parameters.

\paragraph{\Read{}: soft-pool patches into a region token.} A region's token is a learned
weighted average of its patches plus an embedding of the region's size,
\begin{equation}
\begin{aligned}
\alpha_j &= \operatorname{softmax}_{j \in c}\!\big(\mathbf{w}^\top \mathbf{h}^{(s_0)}_{j}\big), \\
\mathbf{z}_c &= \textstyle\sum_{j \in c} \alpha_{j}\, \mathbf{h}^{(s_0)}_{j}
             + \mathbf{e}_{\text{size}}\big(\lfloor \log_2 |c| \rfloor\big) ,
\end{aligned}
\label{eq:read}
\end{equation}
with a single learned score vector $\mathbf{w} \in \mathbb{R}^{d}$. Because the softmax runs
\emph{within} $c$, \Read{} is a region-restricted single-query attention rather than a blind
mean, and can emphasize a region's informative patches. Sizes are arbitrary integers, hence
the index $\lfloor \log_2 |c| \rfloor$ on $\mathbf{e}_{\text{size}}$, telling the core whether
a token stands for one patch or sixteen. The token also carries a \emph{position}, the
average of its patches' rotary embeddings. Since they are unit phasors, the average stays
near the unit circle at wavelengths longer than the region's extent and shrinks toward zero
below it. Averaging low-passes the position in proportion to the region's extent, sharp for a single
patch and coarse for a large one, the right encoding for a token that covers an area rather
than a point.

\paragraph{\Write{}: scatter a region token back to its patches.} Let $\mathbf{z}'_c$ be
region $c$'s token after the core. The naive inverse adds the region's change
$\mathbf{z}'_c - \mathbf{z}_c$ identically to every patch it covers. Instead each patch forms
its own update from that change and its \emph{pre-pool} features,
\begin{equation}
\mathbf{h}^{(s_1+1)}_j = \mathbf{h}^{(s_0)}_{j} +
g\big(\big[\, \mathbf{h}^{(s_0)}_{j} \;\big\|\; \mathbf{z}'_{c(j)} - \mathbf{z}_{c(j)} \,\big]\big),
\label{eq:write}
\end{equation}
where $g$ is a small learned map and $[\,\cdot \,\|\, \cdot\,]$ is concatenation. Patches
in one region can therefore specialize the shared coarse update instead of moving in lockstep. The form is
also a skip around the core: every patch keeps its pre-core features and receives the core's
work as a correction, which is how the coda still produces per-pixel output although the core
never saw individual patches.

\section{Experiments}
\label{sec:experiments}

In \cref{sec:exp-main} we measure what a budget costs in quality and buys in speed, and
compare the region partition against other reductions at matched budgets. In
\cref{sec:exp-ablations} we test whether one checkpoint serves budgets it never trained on,
and what each design choice contributes. In \cref{sec:exp-mechanism} we follow where
the partition spends its fine tokens.

\begin{figure}[t]
\centering
\input{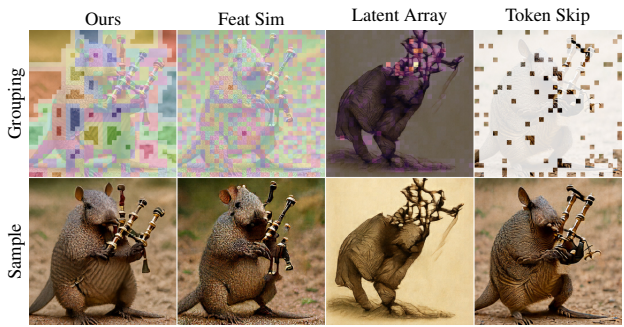}
\caption{\textbf{Reduction methods compared at $\Rcells{=}256$, B/16.} Grouping above, the
resulting sample below. Patches are tinted by the group they pool into; the latent
array forms no groups, so we show one latent's read attention, and Token Skip's dropped
patches are faded. Our regions stay in one place, feature similarity scatters its groups
across the frame, and the latent's read gathers on the salient content, leaving the rest of
the scene unread.}
\label{fig:mechanism}
\end{figure}

\begin{figure*}[!t]
\centering
\input{figures/fig_steps_combined}
\caption{\textbf{Region tokens buy a longer sampling trajectory at matched cost.} On the
MiniT2I-L/16 backbone, we compare \method{} at $\Rcells{=}256$ to shorter dense
trajectories. \textbf{(a)}~Column headings give the time to generate one image on an
RTX~4090; the frozen backbone spends it on fewer steps, \method{} on more. Prompts and seeds
drawn at random from PartiPrompts. \textbf{(b,c)}~GenEval and ImageReward against generation
time. \textit{Please refer to the appendix for the full cost sweep}.}
\label{fig:steps}
\end{figure*}

\begin{figure}[t]
\centering
\centering
\pgfplotsset{
  eastyle/.style={
    scale only axis=true, width=2.90cm, height=1.54cm,
    ylabel style={font=\scriptsize}, xlabel style={font=\scriptsize},
    tick label style={font=\scriptsize}, xlabel shift=-2pt, ylabel shift=-2pt,
    axis line style={line width=0.6pt, draw=gray!70},
    tick style={draw=gray!70, line width=0.6pt}, tick align=inside, minor tick num=0,
    clip=false,
  }}
\begin{minipage}[b]{0.49\linewidth}\centering
{\scriptsize\color{black}
  \textcolor{gblue}{\rule[0.3ex]{10pt}{1.4pt}}\;in-domain\quad
  \textcolor{gred}{$\circ$}\;OOD}
\end{minipage}\hfill
\begin{minipage}[b]{0.49\linewidth}\centering
{\scriptsize\color{black}
  \textcolor{gred}{$\circ$}\kern-0.15em\textcolor{gyellow}{$\circ$}\kern-0.15em\textcolor{ggreen}{$\circ$}\,naive\;\;
  \textcolor{gblue}{$\bullet$}\,\method{} at that row's budget}
\end{minipage}\\[1pt]
\begin{minipage}[t]{0.49\linewidth}\centering
\begin{tikzpicture}
\begin{axis}[eastyle,
    xmode=log, log basis x=2, xmin=56, xmax=590,
    xtick={64,128,256,512}, xticklabels={$64$,$128$,$256$,$512$},
    ymin=70, ymax=90, ytick={72,78,84,90},
    xlabel={\scriptsize\textbf{Budget $\Rcells$}}, ylabel={\scriptsize\textbf{GenEval Acc.\ (\%)}},
    ymajorgrids=true, grid style={line width=0.3pt, draw=gray!30, densely dotted}]
  \addplot[gray!60, densely dashed, line width=0.7pt]
    coordinates {(56,87.2)(590,87.2)};                              %
  \addplot[gblue, line width=1.0pt, mark=*, mark size=1.7pt, mark options={fill=gblue}]
    coordinates {(64,71.8)(128,81.3)(256,84.7)(512,87.8)};        %
  \addplot[only marks, gred, mark=o, mark size=2.1pt, line width=1.0pt]
    coordinates {(96,79.0)(192,84.6)(384,86.4)};                   %
\end{axis}
\end{tikzpicture}\\[2pt]{\footnotesize (a) Elastic budgets.}
\end{minipage}\hfill
\begin{minipage}[t]{0.49\linewidth}\centering
\begin{tikzpicture}
\begin{axis}[eastyle,
    xmin=42, xmax=92, xtick={50,70,90}, xticklabels={$50$,$70$,$90$},
    ytick={1,2,3}, yticklabels={Partition, Interface, Anchoring},
    y tick label style={font=\scriptsize}, ymin=0.35, ymax=3.5,
    xlabel={\scriptsize\textbf{GenEval Acc.\ (\%)}},
    xmajorgrids=true, grid style={line width=0.3pt, draw=gray!30, densely dotted}]
  \draw[gblue!60, densely dotted, line width=0.7pt] (axis cs:81.3,0.76)--(axis cs:81.3,1.24);
  \draw[gblue!60, densely dotted, line width=0.7pt] (axis cs:84.7,1.76)--(axis cs:84.7,2.24);
  \draw[gblue!60, densely dotted, line width=0.7pt] (axis cs:84.7,2.76)--(axis cs:84.7,3.24);
  \draw[gray!55, line width=1.3pt] (axis cs:78.6,1)--(axis cs:81.3,1);
  \draw[gray!55, line width=1.3pt] (axis cs:77.5,2)--(axis cs:84.7,2);
  \draw[gray!55, line width=1.3pt] (axis cs:45.7,3)--(axis cs:84.7,3);
  \addplot[only marks, mark=o, gred, mark size=2.2pt, line width=0.9pt] coordinates {(78.6,1)};
  \addplot[only marks, mark=o, gyellow, mark size=2.2pt, line width=0.9pt] coordinates {(77.5,2)};
  \addplot[only marks, mark=o, ggreen, mark size=2.2pt, line width=0.9pt] coordinates {(45.7,3)};
  \addplot[only marks, mark=*, gblue, mark size=2.4pt, mark options={fill=gblue}]
    coordinates {(81.3,1)(84.7,2)(84.7,3)};
  \node[font=\tiny, text=gred, anchor=south east] at (axis cs:78,0.6) {fixed boundaries};
  \node[font=\tiny, text=gyellow!85!black, anchor=south east] at (axis cs:75.0,1.55) {mean-pool};
  \node[font=\tiny, text=ggreen!90!black, anchor=south east] at (axis cs:65.5,2.2) {latent array};
\end{axis}
\end{tikzpicture}\\[2pt]{\footnotesize (b) Component contribution.}
\end{minipage}
\caption{\textbf{Elasticity and ablations.} \textbf{(a)}~One \method{} checkpoint read
across the budget range; the four training budgets and three it never saw (red) fall on one
curve, with the frozen base dashed. \textbf{(b)}~One design choice swapped at a
time, named in the panel; the open dot marks where quality lands. The partition swap runs
at $\Rcells{=}128$, where boundary placement matters most, the others at $\Rcells{=}256$.}
\label{fig:elastic_ablation}
\end{figure}
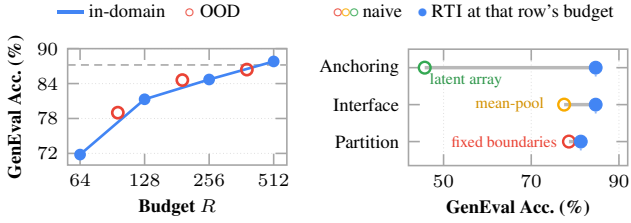

\begin{figure*}[t]
\centering
\input{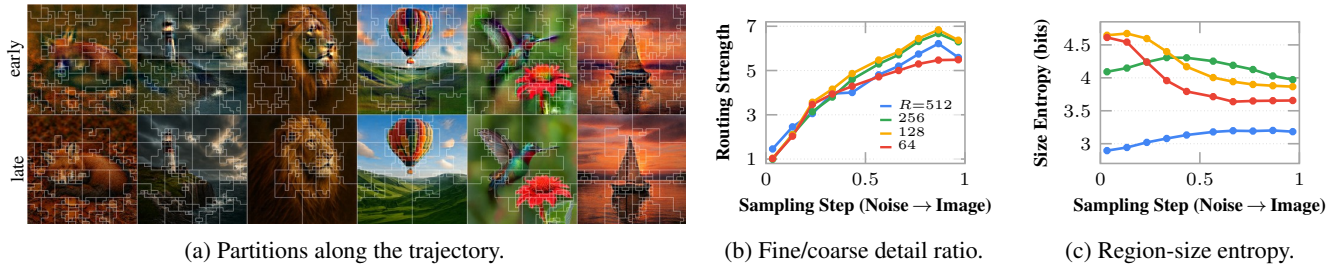}
\caption{\textbf{The partition tracks the image as it forms.} One elastic \method{} checkpoint,
(a) at $\Rcells{=}128$ and (b,c) at four budgets. \textbf{(a)}~Six prompts early and late in
sampling, the image estimate with region boundaries overlaid; the tiling stays coarse over
flat areas and refines onto the objects as they resolve. \textbf{(b)}~The ratio of detail in
the fine regions to that in the coarse ones, near $1$ at pure noise and rising to
${\sim}6\times$ as detail gathers. \textbf{(c)}~The region-size entropy over sampling,
rising from near-uniform at $\Rcells{=}512$ and falling from strongly varied at
$\Rcells{=}64$. \textit{Please refer to the appendix for more details}.}
\label{fig:routing}
\end{figure*}

\subsection{Setup}
\label{sec:exp-setup}

\paragraph{Model and training.} We start from the pretrained MiniT2I-B/16~\cite{minit2i2026}, keep it frozen, and train only the LoRA adapters and the
\Read{}/\Write{} interface, initialised to the identity so that training starts from the
frozen model's behaviour. The core spans blocks $[3,13]$ of $17$, leaving a three-block
prelude and coda, and the budget $\Rcells$ is drawn for each training step from
$\{64,128,256,512\}$ so that one checkpoint covers the range. All other settings follow
MiniT2I's fine-tuning recipe, $40$K steps on its ${\sim}120$K-image alignment pool at
$512^2$, outlined in the appendix. On L/16 the same recipe compresses $[3,19]$
of $23$ blocks.
Every baseline in \cref{sec:exp-main} is our reimplementation, trained with the same recipe
on the same backbone.

\paragraph{Metrics.} We report GenEval overall accuracy over its $553$
prompts~\citep{ghoshGenEvalObjectFocusedFramework2023} and three preference scores over all
$1632$ PartiPrompts~\citep{yu2022parti}: CLIPScore~\citep{hessel2021clipscore},
PickScore~\citep{kirstain2023pickapic} and ImageReward~\citep{xu2023imagereward}. Sampling is
$100$ Euler steps at guidance $5.0$ following the baseline settings, at $\Rcells{=}256$ unless stated otherwise.

\paragraph{Speed.} We time the full generation loop on an RTX~4090 with fused attention and
four images per batch, doubled to eight by classifier-free guidance, and report the median
images per second over three interleaved passes. The timing includes everything the
interface adds, the partition, \Read{}/\Write{}, the adapters and the text stream; a speedup
is the ratio of two throughputs on the same backbone.

\subsection{Main Results}
\label{sec:exp-main}

At $\Rcells{=}256$, \method{} runs the frozen B/16 backbone at $2.15\times$ throughput for
$84.7$ GenEval against the dense $87.2$. Sweeping the budget of that same checkpoint traces
the quality--speed curve of \cref{fig:frontier}, where the difference falls inside noise
from $\Rcells{=}384$ upward, at $2.00\times$, though dense stays ahead on every preference
score. Speed saturates near $2.5\times$ when only the prelude, coda and text stream remain,
so below $\Rcells{\approx}192$ a leaner budget costs quality without buying speed.

On B/16 \method{} scores highest overall and is matched or ahead on five of six GenEval
categories. At $\Rcells{=}512$ it clears its own dense backbone on quality and speed at
once, $87.8$ against $87.2$ at $1.84\times$, while Feat Sim at the same budget and speed
only draws level with our $\Rcells{=}256$ row. On L/16, whose longer trunk leaves more to
compress, $\Rcells{=}512$ holds $87.5$ within noise of dense at $2.11\times$; the block
shows transfer to a longer trunk, not a case for choosing L/16 over B/16.

\paragraph{Comparison to other reduction methods.} We compare against three families, each giving up at
least one of the four properties of \cref{sec:intro} --- contiguity, content-sized regions, position,
summarization --- and each implemented inside our own interface so that only the reduction
differs. \cref{fig:mechanism} pairs what each does to the tokens with the resulting sample,
and we provide an analysis in the appendix which measures each grouping's spatial spread,
the mean distance from a group's patches to its centre in units of the patch grid, and the
share of the span representation its pooling retains; we quote both below. All three also forfeit content-sized regions and differ
in the remaining properties.

\emph{Token skipping} lets tokens bypass blocks rather than pooling them. We follow
Mixture-of-Depths~\citep{raposo2024mod}, whose core blocks alternate dense and sparse; a
sparse block admits the top $12.5\%$ of image tokens, and the rest pass on the residual.
What it gives up is summarization, retaining a third of what our grouping keeps.
It is the strongest baseline in the $\Rcells{=}256$ comparison, $82.3$ GenEval, yet also
the slowest, $1.19\times$. Both follow from the dense half of the core, which runs in full
at any capacity, preserving quality but capping the speedup below ${\sim}1.3\times$, short
of iso-compute with our framework.

\emph{Feature-similarity merging}, as in ToMe~\citep{bolya2023tome} and its diffusion
adaptation~\citep{bolya2023tomesd}, groups by similarity irrespective of location, so one
group may draw patches from across the frame (\cref{fig:mechanism}); its patches sit $5.2$
grid cells from their group's centre, ours $0.5$. What it loses is contiguity, and with it position. Information is kept, the merge retaining
as much of the span representation as our region tokens.
The cost grows as the budget thins, reaching $13.8$ GenEval points at $\Rcells{=}64$, and at
$\Rcells{=}256$ it trails us on every preference score on both backbones.

A \emph{latent array}, as in Perceiver-IO and RIN~\citep{jaegle2022perceiverio,jabri2023rin},
forgoes grouping altogether, and with it contiguity and position at once: $\Rcells$ learned
latents read the whole sequence by cross-attention and write it back the same way. After $40$K steps a latent's read attention still covers essentially the whole frame
(\cref{fig:mechanism}), a spread of $11.4$ grid cells of a possible $12.2$. What structure the read does have is semantic rather than spatial. The latents gather on salient content, leaving the rest of a busy scene read by no latent
in particular; content a partition covers by construction goes unrepresented. Generation is poor, at $45.7$ GenEval. Perceiver-IO and RIN were designed for training from scratch; we measure whether a latent
bottleneck can inherit a frozen backbone.

Each baseline is worst exactly where its design gives a property up. Against \method{} at
the same budget, the costs are $2.4$ GenEval points for deleting rather than summarizing,
$4.2$ for scattering the group, and $39.0$ for removing the position; the first understates
its gap, since Token Skip runs at $1.19\times$ to our $2.15\times$. For class-conditional
generation with a pixel-space JiT~\citep{liBackBasicsLet2026}, please refer to the appendix.

\paragraph{Visual results.} \cref{fig:gallery} sets \method{} against feature-similarity
merging, budget for budget on the same prompts and seeds. Both hold the dense image at
$\Rcells{=}512$; at $\Rcells{=}256$, a quarter of the tokens and $2.59\times$ the
throughput, \method{} still preserves the layout and fine detail of the dense backbone while the merge
comes apart, lettering first and structure with it.
\cref{fig:mechanism} traces the failure to the tokens: merging blends features from
different parts of the image, the signature of lost positions, and the latent array,
anchored nowhere, loses the subject.

\paragraph{Fewer sampling steps.} Fewer steps and a smaller region budget save the same
wall-clock, so the two are alternatives; \cref{fig:steps} compares them at equal cost. A step on
region tokens is $2.6\times$ cheaper, so the same budget buys a $2.6\times$ longer
trajectory. At $1.8$ seconds the dense backbone affords $6$ steps and cannot resolve an
image, while our $16$ are already clean (\cref{fig:steps}a), a gap of $16.8$ GenEval points
(\cref{fig:steps}b). The advantage persists at $2.8$ seconds (\cref{fig:steps}c)
and reverses above roughly five seconds an image, where more steps become the better
purchase; the $100$-step setting of \cref{tab:main} thus matches dense quality, but is not
the cheapest way to reach it.

\subsection{Elasticity and Ablations}
\label{sec:exp-ablations}

\paragraph{Elasticity.} The curve of \cref{fig:frontier} is read at the four budgets seen
during training. Evaluated at three it never saw ($\Rcells\in\{96,192,384\}$), the same
checkpoint lands on the same curve (\cref{fig:elastic_ablation}a): within this range the
budget is a continuous knob, and can be set after training.

\paragraph{Contribution of each component.} We swap one design choice at a time and measure
what that costs (\cref{fig:elastic_ablation}b). The partition swap spaces the cuts
evenly along the Hilbert order instead of placing them at the largest feature gaps, holding
the ordering, the budget and the interface fixed. It measures adaptive against fixed
boundaries, not region tokens against patch tokens, and costs $2.7$ GenEval points at
$\Rcells{=}128$, less as the budget grows, since at
$\Rcells{=}256$ a region holds four patches and an evenly spaced cut already lands close to
the adaptive one. Replacing the learned \Read{}/\Write{} with a mean and a broadcast costs $7.2$. Replacing
the anchored region tokens with a latent array drops quality to the Latent Array baseline of
\cref{tab:main}, since the rotary attention then receives no position at all. This swap
removes grouping together with position, so its cost is an upper bound on the contribution
of anchoring alone.

\subsection{Where the Partition Spends Its Tokens}
\label{sec:exp-mechanism}

We investigate how the partitions behave across the sampling trajectory. In
\cref{fig:routing}a the single-patch regions gather on the objects and their boundaries;
flat sky, water and background are covered by a few large ones. To
quantify this, we take a patch's detail to be the gradient energy of the clean estimate
$\hat x$, and call a region fine at one patch and coarse at sixteen or more. In
\cref{fig:routing}b the two carry the same detail at the noisiest steps and pull apart to a
ratio of roughly six as the image forms. This alignment is not built in. The cut reads the features entering the core and
never the detail measure, so the agreement between fine regions and image detail is the model's doing. The image does
not decide how many scales the tiling uses. A rich budget lets most regions stay
small; a lean one covers the same frame with few tokens, so single patches on detail coexist
with large flat regions. \cref{fig:routing}c shows this as the
region-size entropy, low at $\Rcells{=}512$, high at $\Rcells{=}64$, drifting toward a
varied middle over sampling. The budget sets
the range of sizes, and the image places the fine regions inside it. For a more detailed
analysis, please refer to the appendix.

\section{Conclusion and Limitations}
\label{sec:discussion}

Compressing tokens raises three questions: whether they are redundant, where the redundancy
sits, and what a reduction must preserve. Probing a frozen pixel transformer answered the
first two and turned the third into four requirements, fixed before any reduction was
trained; they said in advance which reduction would fail and how. The comparisons in
\cref{sec:exp-main} confirm all three predictions.

A Hilbert curve meets all four requirements at once. What is new is not a mechanism but the
observation that regions rather than patches can be the token unit of a model already
trained on patches. The interface is elastic, one checkpoint holding every budget;
the operating point is set on deployment, not in training.

\paragraph{Limitations.} The speedup has a ceiling, since prelude, coda and text stream
never shrink. The core cannot distinguish patches within a region. Not every grouping is a Hilbert
interval, and the budget is chosen by the user, not by the image content.

\section*{Acknowledgments}
This work was supported by the Alexander von Humboldt Foundation.

\bibliography{main_lib,refs_extra}
\fi

\ifdefined\appendixonly
  \setcounter{figure}{9}\setcounter{table}{1}\setcounter{equation}{4}
  \appendix
  \twocolumn[%
  \begin{center}
    {\LARGE\bf \papertitle\par}\vspace{7pt}
    {\Large Technical Appendix\par}
  \end{center}
  \vspace{14pt}
]

\setcounter{topnumber}{2}\setcounter{dbltopnumber}{2}\setcounter{totalnumber}{3}
\renewcommand{\topfraction}{0.85}\renewcommand{\dbltopfraction}{0.78}
\renewcommand{\textfraction}{0.10}
\renewcommand{\floatpagefraction}{0.75}\renewcommand{\dblfloatpagefraction}{0.75}

\Cref{sec:supp-impl} lists the training and sampling settings behind all reported
numbers, the baseline implementations, and the compute accounting. \Cref{sec:supp-exp}
collects additional experiments: FID on MJHQ-30K, class-conditional generation with JiT,
fewer sampling steps, the partition order, and a longer core. \Cref{sec:supp-probes} gives the
protocols behind the redundancy, grouping and routing analyses; \cref{sec:supp-visuals}
closes with further qualitative results.

\begin{table}[t]
\centering\footnotesize
\setlength{\tabcolsep}{2.5pt}
\begin{tabular}{@{}lccc@{}}
\toprule
 & MiniT2I-B/16 & MiniT2I-L/16 & JiT-B/16 \\
 & \multicolumn{2}{c}{\emph{text-to-image}} & \emph{class-cond.} \\
\midrule
resolution           & $512^2$ & $512^2$ & $256^2$ \\
patch tokens $\Ntok$ & $1024$ & $1024$ & $256$ \\
trunk blocks         & $17$ & $23$ & $12$ \\
core blocks          & $3$--$13$ & $3$--$19$ & $4$--$9$ \\
budgets trained      & \multicolumn{2}{c}{$\{64,128,256,512\}$} & \shortstack{$\{32,64,$\\$128,192\}$} \\
steps                & $40$K & $40$K & $100$K \\
trainable            & $16.3$M & $36.0$M & $6.0$M \\
of the backbone      & $5.9\%$ & $3.8\%$ & $4.4\%$ \\
\midrule
optimizer            & \multicolumn{3}{c}{AdamW, lr $10^{-4}$, $2$K warmup} \\
global batch         & \multicolumn{3}{c}{$256$} \\
LoRA rank            & \multicolumn{3}{c}{$32$} \\
EMA                  & \multicolumn{3}{c}{$0.9999$} \\
\midrule
sampler              & \multicolumn{2}{c}{$100$ Euler steps} & $50$ Heun steps \\
guidance             & \multicolumn{2}{c}{$5.0$} & $3.0$ on $[0.1,1.0]$ \\
\bottomrule
\end{tabular}
\caption{\textbf{Training and sampling settings.} Upper block per backbone, lower
blocks shared by all runs. The text-to-image models train on MiniT2I's ${\sim}120$K-image
alignment pool, the class-conditional models on ImageNet-1K~\citep{deng2009imagenet};
JiT-L/16 (\cref{fig:fid}b,c) follows the JiT-B/16 column with core $8$--$19$ of $24$
blocks. A checkpoint also serves budgets it was not trained on (\cref{sec:exp-ablations});
\cref{fig:fid} reads $\Rcells{\in}\{64,130,193\}$. All baselines share this recipe.}
\label{tab:settings}
\end{table}

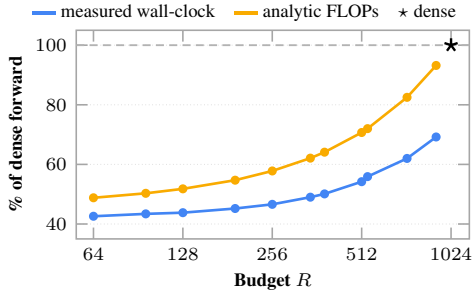
\begin{figure}[tb]
\centering
\centering
{\scriptsize\color{black}
  \textcolor{gblue}{\rule[0.35ex]{9pt}{1.4pt}}\;measured wall-clock\quad
  \textcolor{gyellow}{\rule[0.35ex]{9pt}{1.4pt}}\;analytic FLOPs\quad
  $\star$\;dense}\\[2pt]
\begin{tikzpicture}
\begin{axis}[
    scale only axis=true, width=5.2cm, height=2.76cm,
    xmode=log, log basis x=2, xmin=56, xmax=1180,
    xtick={64,128,256,512,1024}, xticklabels={$64$,$128$,$256$,$512$,$1024$},
    ymin=35, ymax=105, ytick={40,60,80,100},
    xlabel={\scriptsize\textbf{Budget $\Rcells$}},
    ylabel={\scriptsize\textbf{\% of dense forward}},
    ylabel style={font=\scriptsize}, xlabel style={font=\scriptsize},
    tick label style={font=\scriptsize}, xlabel shift=-2pt, ylabel shift=-2pt,
    ymajorgrids=true, grid style={line width=0.3pt, draw=gray!30, densely dotted},
    axis line style={line width=0.6pt, draw=gray!70},
    tick style={draw=gray!70, line width=0.6pt}, tick align=inside, minor tick num=0,
    clip=false]
  \addplot[gray!60, densely dashed, line width=0.7pt] coordinates {(56,100)(1180,100)};
  \addplot[gyellow, line width=1.0pt, mark=*, mark size=1.2pt, mark options={fill=gyellow}]
    coordinates {(64,48.8)(96,50.3)(128,51.8)(192,54.7)(256,57.8)(344,62.1)(384,64.1)
                 (512,70.7)(536,72.0)(728,82.5)(912,93.2)};
  \addplot[gblue, line width=1.0pt, mark=*, mark size=1.2pt, mark options={fill=gblue}]
    coordinates {(64,42.6)(96,43.4)(128,43.8)(192,45.2)(256,46.6)(344,49.0)(384,50.1)
                 (512,54.2)(536,55.9)(728,62.0)(912,69.2)};
  \addplot[black, only marks, mark=star, mark size=2.6pt, line width=0.9pt]
    coordinates {(1024,100)};
\end{axis}
\end{tikzpicture}
\caption{\textbf{Cost of one forward pass against the budget} on B/16, relative to the
dense forward. Measured time (RTX~4090, batch~$4$) falls faster than analytic FLOPs:
attention at these lengths is partly memory-bound.}
\label{fig:flops}
\end{figure}

\begin{figure*}[t]
\centering
\pgfplotsset{fidax/.style={
    width=\linewidth, height=0.70\linewidth,
    ylabel style={font=\scriptsize}, xlabel style={font=\scriptsize},
    tick label style={font=\scriptsize}, xlabel shift=-2pt, ylabel shift=-2pt,
    ymajorgrids=true, grid style={line width=0.3pt, draw=gray!30, densely dotted},
    axis line style={line width=0.6pt, draw=gray!70},
    tick style={draw=gray!70, line width=0.6pt}, tick align=inside, minor tick num=0,
    scaled y ticks=false, clip=false,
}}
\centering
{\scriptsize\color{black}
  \textcolor{gblue}{\rule[0.35ex]{10pt}{1.6pt}}\;\method{} (ours)\qquad
  \textcolor{ggreen}{\rule[0.35ex]{10pt}{1.6pt}}\;Feat Sim\qquad
  \textcolor{gyellow}{$\blacktriangle$}\;Token Skip\qquad
  $\star$\;dense\qquad
  {\color{gray!90}full = B/16, faded = L/16 (b,c)}}\\[1pt]
\begin{minipage}[t]{0.33\textwidth}
\centering
\begin{tikzpicture}
  \begin{axis}[fidax, xmin=0.85, xmax=2.85, xtick={1.0,1.5,2.0,2.5},
    xticklabels={$1\times$,$1.5\times$,$2\times$,$2.5\times$},
    ymin=6, ymax=30.5, ytick={9.07,15,20,25,30}, yticklabels={$9.1$,$15$,$20$,$25$,$30$},
    xlabel={\textbf{Speedup (measured)}}, ylabel={\textbf{FID\,$\downarrow$}}]
  \addplot[ggreen, line width=1.1pt, mark=diamond*, mark size=2.0pt,
           mark options={fill=ggreen}] coordinates {(2.59,28.02)(2.11,19.99)};
  \addplot[gblue, line width=1.3pt, mark=*, mark size=1.5pt, mark options={fill=gblue}]
    coordinates {(2.59,14.71)(2.11,11.43)};
  \addplot[black, only marks, mark=star, mark size=2.8pt, line width=0.9pt]
    coordinates {(1.0,9.07)};
  \draw[-{Stealth[length=3.5pt]}, gred, line width=0.8pt, shorten <=2.5pt, shorten >=2.5pt]
    (axis cs:2.11,19.99) -- (axis cs:2.11,11.43)
    node[midway, anchor=west, font=\scriptsize\bfseries, gred, inner sep=2pt] {$8.6$};
  \draw[-{Stealth[length=3.5pt]}, gred, line width=0.8pt, shorten <=2.5pt, shorten >=2.5pt]
    (axis cs:2.59,28.02) -- (axis cs:2.59,14.71)
    node[midway, anchor=east, font=\scriptsize\bfseries, gred, inner sep=2pt] {$13.3$};
  \node[font=\tiny, gblue, anchor=north] at (axis cs:2.59,13.9) {$\Rcells{=}256$};
  \node[font=\tiny, gblue, anchor=north west] at (axis cs:2.02,10.6) {$512$};
  \end{axis}
\end{tikzpicture}
\par\smallskip
{(a) MJHQ-30K, MiniT2I-L/16.}
\end{minipage}\hfill
\begin{minipage}[t]{0.33\textwidth}
\centering
\begin{tikzpicture}
  \begin{axis}[fidax, xmin=0.94, xmax=1.72, xtick={1.0,1.2,1.4,1.6},
    xticklabels={$1\times$,$1.2\times$,$1.4\times$,$1.6\times$},
    ymode=log, ymin=2.05, ymax=30, ytick={2.5,5,10,20},
    yticklabels={$2.5$,$5$,$10$,$20$},
    xlabel={\textbf{Speedup (token compute)}}, ylabel={\textbf{FID\,$\downarrow$}}]
  \addplot[ggreen, line width=1.1pt, mark=diamond*, mark size=2.0pt,
           mark options={fill=ggreen}] coordinates {(1.613,23.49)(1.333,15.99)(1.136,7.26)};
  \addplot[gblue, line width=1.3pt, mark=*, mark size=1.5pt, mark options={fill=gblue}]
    coordinates {(1.613,16.69)(1.333,6.23)(1.136,4.32)};
  \addplot[gyellow, densely dashed, line width=1.0pt, mark=triangle*, mark size=2.5pt,
           mark options={fill=gyellow, solid}] coordinates {(1.408,19.11)(1.0,7.02)};
  \addplot[black, only marks, mark=star, mark size=2.8pt, line width=0.9pt]
    coordinates {(1.0,3.62)};
  \addplot[ggreen, opacity=0.45, line width=1.1pt, mark=diamond, mark size=2.2pt]
    coordinates {(1.613,6.28)(1.333,4.53)(1.136,2.73)};
  \addplot[gblue, opacity=0.45, line width=1.3pt, mark=square*, mark size=1.5pt]
    coordinates {(1.613,4.71)(1.333,2.69)(1.136,2.42)};
  \addplot[black, opacity=0.45, only marks, mark=star, mark size=2.8pt, line width=0.9pt]
    coordinates {(1.0,2.36)};
  \end{axis}
\end{tikzpicture}
\par\smallskip
{(b) ImageNet-$256$, JiT.}
\end{minipage}\hfill
\begin{minipage}[t]{0.33\textwidth}
\centering
\begin{tikzpicture}
  \begin{axis}[fidax, xmin=0.94, xmax=1.72, xtick={1.0,1.2,1.4,1.6},
    xticklabels={$1\times$,$1.2\times$,$1.4\times$,$1.6\times$},
    ymin=95, ymax=335, ytick={100,150,200,250,300},
    yticklabels={$100$,$150$,$200$,$250$,$300$},
    xlabel={\textbf{Speedup (token compute)}},
    ylabel={\textbf{Inception Score\,$\uparrow$}}]
  \addplot[ggreen, line width=1.1pt, mark=diamond*, mark size=2.0pt,
           mark options={fill=ggreen}] coordinates {(1.613,109.0)(1.333,137.2)(1.136,199.1)};
  \addplot[gblue, line width=1.3pt, mark=*, mark size=1.5pt, mark options={fill=gblue}]
    coordinates {(1.613,136.1)(1.333,216.1)(1.136,241.1)};
  \addplot[gyellow, densely dashed, line width=1.0pt, mark=triangle*, mark size=2.5pt,
           mark options={fill=gyellow, solid}] coordinates {(1.408,121.2)(1.0,200.6)};
  \addplot[black, opacity=0.45, only marks, mark=star, mark size=2.8pt, line width=0.9pt]
    coordinates {(1.0,275.1)};
  \addplot[ggreen, opacity=0.45, line width=1.1pt, mark=diamond, mark size=2.2pt]
    coordinates {(1.613,229.2)(1.333,252.8)(1.136,291.6)};
  \addplot[gblue, opacity=0.45, line width=1.3pt, mark=square*, mark size=1.5pt]
    coordinates {(1.613,253.0)(1.333,303.8)(1.136,313.5)};
  \addplot[black, opacity=0.45, only marks, mark=star, mark size=2.8pt, line width=0.9pt]
    coordinates {(1.0,298.5)};
  \end{axis}
\end{tikzpicture}
\par\smallskip
{(c) ImageNet-$256$, JiT.}
\end{minipage}
\caption{\textbf{Distribution-level quality.} Every panel plots quality against speedup
over the dense model: dense sits at $1\times$, compression moves right.
\textbf{(a)}~$30$K prompts, clean-fid; the speedup is measured wall-clock. \method{}
and Feat Sim run at the same speed at a given budget, so they share an $x$ position and
the \textcolor{gred}{red arrows} give the cost of the grouping rule alone.
\textbf{(b,c)}~$50$K samples per measured point,
torch-fidelity~\citep{obukhov2020torchfidelity};
the speedup is the analytic one implied by token compute, as these backbones were not
timed. $\Rcells{=}64/130/193$ gives
$1.61/1.33/1.14\times$ on both backbones, which compress half their trunk. The dashed
line joins the two operating points of Token Skip: as trained it runs at $1.41\times$;
returning its saving as three extra depth groups brings it back to dense compute but
not to dense quality ($7.02$ against $3.62$ FID). Full colour is JiT-B/16, faded JiT-L/16.
\method{} leads Feat Sim at every budget in all three panels.}
\label{fig:fid}
\end{figure*}
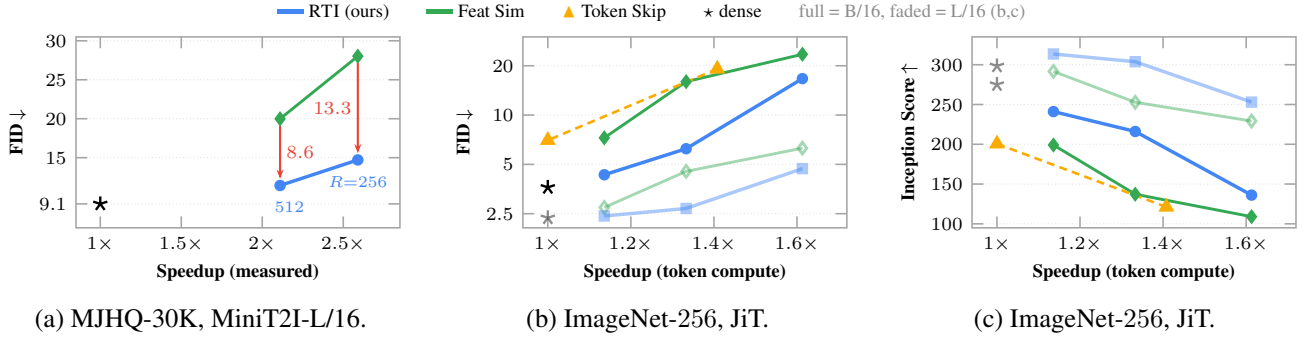

\begin{figure}[t]
\centering
\pgfplotsset{pooljitax/.style={
    scale only axis=true, width=2.9cm, height=2.5cm,
    ylabel style={font=\scriptsize}, xlabel style={font=\scriptsize},
    tick label style={font=\scriptsize}, xlabel shift=-2pt, ylabel shift=-2pt,
    ymajorgrids=true, grid style={line width=0.3pt, draw=gray!30, densely dotted},
    axis line style={line width=0.6pt, draw=gray!70},
    tick style={draw=gray!70, line width=0.6pt}, tick align=inside, minor tick num=0,
    xmin=-0.02, xmax=1.02, xtick={0,0.5,1}, xticklabels={0,0.5,1},
    ymin=0.585, ymax=0.80, ytick={0.60,0.65,0.70,0.75,0.80},
    yticklabels={60,65,70,75,80},
    xlabel={\textbf{Relative Depth}},
    clip=false,
}}
\centering
{\scriptsize\color{black}
  \textcolor{gblue}{\rule[0.35ex]{9pt}{1.6pt}}\;B/16\quad
  \textcolor{gblue!45}{\rule[0.35ex]{9pt}{1.6pt}}\;L/16\quad
  \textcolor{gred}{\rule[0.35ex]{9pt}{1.6pt}}\;B/32\quad
  \textcolor{gred!45}{\rule[0.35ex]{9pt}{1.6pt}}\;L/32\quad
  \textcolor{gpurple}{\rule[0.35ex]{9pt}{1.6pt}}\;H/32\quad
  \textcolor{ggreyd}{\rule[0.35ex]{9pt}{1.0pt}}\;MiniT2I-B (t2i)}\\[2pt]
\begin{tikzpicture}
\begin{groupplot}[group style={group size=2 by 1, horizontal sep=30pt}, pooljitax]
\nextgroupplot[ylabel={\textbf{Repr.\ Retained (\%)}},
    title={\scriptsize\textbf{$256^2$ ($16^2$ px/patch)}}, title style={yshift=-3pt}]
  \addplot[gblue, line width=1.0pt, mark=*, mark size=0.9pt, mark options={fill=gblue}]
    coordinates {
    (0.000,0.701)(0.091,0.713)(0.182,0.733)(0.273,0.733)(0.364,0.731)(0.455,0.726)
    (0.545,0.722)(0.636,0.722)(0.727,0.722)(0.818,0.707)(0.909,0.703)(1.000,0.73)};
  \addplot[gblue, opacity=0.45, line width=1.0pt, mark=square*, mark size=0.9pt]
    coordinates {
    (0.000,0.716)(0.043,0.734)(0.087,0.734)(0.130,0.741)(0.174,0.738)(0.217,0.738)
    (0.261,0.735)(0.304,0.74)(0.348,0.742)(0.391,0.743)(0.435,0.745)(0.478,0.746)
    (0.522,0.743)(0.565,0.742)(0.609,0.741)(0.652,0.737)(0.696,0.732)(0.739,0.731)
    (0.783,0.723)(0.826,0.712)(0.870,0.695)(0.913,0.694)(0.957,0.687)(1.000,0.738)};
\nextgroupplot[title={\scriptsize\textbf{$512^2$ ($32^2$ px/patch)}}, title style={yshift=-3pt}]
  \addplot[ggreyd, densely dashed, line width=0.9pt]
    coordinates {
    (0.000,0.614)(0.062,0.675)(0.125,0.713)(0.188,0.729)(0.250,0.735)(0.312,0.739)
    (0.375,0.74)(0.438,0.741)(0.500,0.749)(0.562,0.747)(0.625,0.74)(0.688,0.732)
    (0.750,0.723)(0.812,0.721)(0.875,0.717)(0.938,0.688)(1.000,0.719)};
  \addplot[gred, line width=1.0pt, mark=*, mark size=0.9pt, mark options={fill=gred}]
    coordinates {
    (0.000,0.613)(0.091,0.692)(0.182,0.694)(0.273,0.69)(0.364,0.688)(0.455,0.688)
    (0.545,0.691)(0.636,0.702)(0.727,0.698)(0.818,0.686)(0.909,0.665)(1.000,0.655)};
  \addplot[gred, opacity=0.45, line width=1.0pt, mark=square*, mark size=0.9pt]
    coordinates {
    (0.000,0.609)(0.043,0.654)(0.087,0.665)(0.130,0.688)(0.174,0.695)(0.217,0.699)
    (0.261,0.699)(0.304,0.708)(0.348,0.709)(0.391,0.71)(0.435,0.713)(0.478,0.718)
    (0.522,0.722)(0.565,0.72)(0.609,0.725)(0.652,0.725)(0.696,0.735)(0.739,0.735)
    (0.783,0.739)(0.826,0.743)(0.870,0.729)(0.913,0.729)(0.957,0.726)(1.000,0.723)};
  \addplot[gpurple, line width=1.0pt, mark=triangle*, mark size=1.2pt,
           mark options={fill=gpurple}]
    coordinates {
    (0.000,0.676)(0.032,0.695)(0.065,0.7)(0.097,0.695)(0.129,0.699)(0.161,0.712)
    (0.194,0.712)(0.226,0.714)(0.258,0.715)(0.290,0.718)(0.323,0.718)(0.355,0.719)
    (0.387,0.721)(0.419,0.724)(0.452,0.725)(0.484,0.73)(0.516,0.733)(0.548,0.739)
    (0.581,0.746)(0.613,0.752)(0.645,0.757)(0.677,0.763)(0.710,0.767)(0.742,0.773)
    (0.774,0.775)(0.806,0.777)(0.839,0.779)(0.871,0.769)(0.903,0.753)(0.935,0.757)
    (0.968,0.742)(1.000,0.739)};
\end{groupplot}
\end{tikzpicture}
\caption{\textbf{The redundancy probe of \cref{fig:poolability}c on the frozen JiT
family}, split by resolution. Each curve: representation retained by the adaptive
partition at $25\%$ of the patch tokens, against relative depth; $64$ classes per model,
each backbone's own sampler, partition cut at the core start. Dashed: the t2i backbone
of \cref{fig:poolability}c. The level of redundancy is similar everywhere
($0.61$--$0.78$); what changes is the depth profile: the hourglass is shallow at $256^2$ and deepens
with resolution and model size (numbers in the text).}
\label{fig:poolability-jit}
\end{figure}
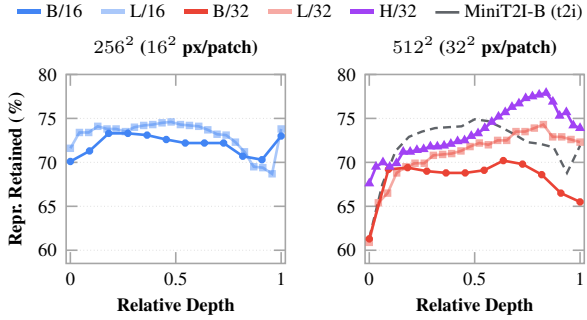

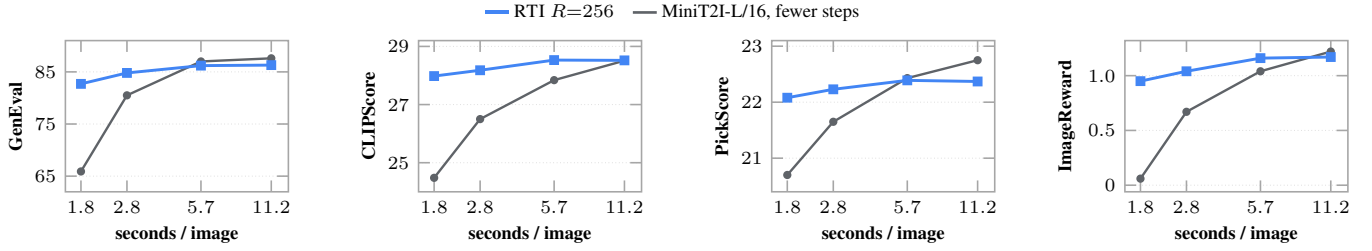
\begin{figure*}[t]
\centering
\centering
{\scriptsize\color{black}
  \textcolor{gblue}{\rule[0.35ex]{9pt}{1.4pt}}\;\method{} $\Rcells{=}256$\quad
  \textcolor{ggreyd}{\rule[0.35ex]{9pt}{1.0pt}}\;MiniT2I-L/16, fewer steps}\\[3pt]
\begin{tikzpicture}
\begin{groupplot}[
    group style={group size=4 by 1, horizontal sep=49pt},
    scale only axis=true, width=2.95cm, height=2.0cm,
    xmode=log, xmin=1.55, xmax=13.2,
    xtick={1.8,2.8,5.7,11.2}, xticklabels={$1.8$,$2.8$,$5.7$,$11.2$},
    xlabel={\scriptsize\textbf{seconds / image}},
    ylabel style={font=\scriptsize}, xlabel style={font=\scriptsize},
    tick label style={font=\scriptsize}, xlabel shift=-2pt, ylabel shift=-2pt,
    ymajorgrids=true, grid style={line width=0.3pt, draw=gray!30, densely dotted},
    axis line style={line width=0.6pt, draw=gray!70},
    tick style={draw=gray!70, line width=0.6pt}, tick align=inside, minor tick num=0,
    clip=false]
\nextgroupplot[ymin=62, ymax=91, ytick={65,75,85},
    ylabel={\scriptsize\textbf{GenEval}}]
  \addplot[ggreyd, line width=0.9pt, mark=*, mark size=1.1pt, mark options={fill=ggreyd}]
    coordinates {(1.8,65.9)(2.8,80.5)(5.7,87.0)(11.2,87.6)};
  \addplot[gblue, line width=1.1pt, mark=square*, mark size=1.3pt, mark options={fill=gblue}]
    coordinates {(1.8,82.7)(2.8,84.8)(5.7,86.2)(11.2,86.3)};
\nextgroupplot[ymin=24, ymax=29.2, ytick={25,27,29},
    ylabel={\scriptsize\textbf{CLIPScore}}]
  \addplot[ggreyd, line width=0.9pt, mark=*, mark size=1.1pt, mark options={fill=ggreyd}]
    coordinates {(1.8,24.48)(2.8,26.50)(5.7,27.84)(11.2,28.51)};
  \addplot[gblue, line width=1.1pt, mark=square*, mark size=1.3pt, mark options={fill=gblue}]
    coordinates {(1.8,27.98)(2.8,28.18)(5.7,28.53)(11.2,28.52)};
\nextgroupplot[ymin=20.4, ymax=23.1, ytick={21,22,23},
    ylabel={\scriptsize\textbf{PickScore}}]
  \addplot[ggreyd, line width=0.9pt, mark=*, mark size=1.1pt, mark options={fill=ggreyd}]
    coordinates {(1.8,20.70)(2.8,21.65)(5.7,22.43)(11.2,22.75)};
  \addplot[gblue, line width=1.1pt, mark=square*, mark size=1.3pt, mark options={fill=gblue}]
    coordinates {(1.8,22.08)(2.8,22.23)(5.7,22.39)(11.2,22.37)};
\nextgroupplot[ymin=-0.06, ymax=1.32, ytick={0,0.5,1.0}, yticklabels={$0$,$0.5$,$1.0$},
    ylabel={\scriptsize\textbf{ImageReward}}]
  \addplot[ggreyd, line width=0.9pt, mark=*, mark size=1.1pt, mark options={fill=ggreyd}]
    coordinates {(1.8,0.06)(2.8,0.67)(5.7,1.04)(11.2,1.22)};
  \addplot[gblue, line width=1.1pt, mark=square*, mark size=1.3pt, mark options={fill=gblue}]
    coordinates {(1.8,0.95)(2.8,1.04)(5.7,1.16)(11.2,1.17)};
\end{groupplot}
\end{tikzpicture}
\caption{\textbf{The full cost sweep behind \cref{fig:steps}: region compression against
fewer sampling steps on MiniT2I-L/16}, all four metrics against wall-clock time per
image. At the four matched costs ($1.8$/$2.8$/$5.7$/$11.2$\,s, within $4\%$ per pair) the
dense backbone runs $6$/$10$/$20$/$39$ steps and \method{} at $\Rcells{=}256$ runs
$16$/$25$/$50$/$100$. \method{} leads everywhere at the two low costs, splits at $5.7$\,s, trails at
$11.2$\,s; on CLIPScore the dense model never catches up.}
\label{fig:steps-supp}
\end{figure*}

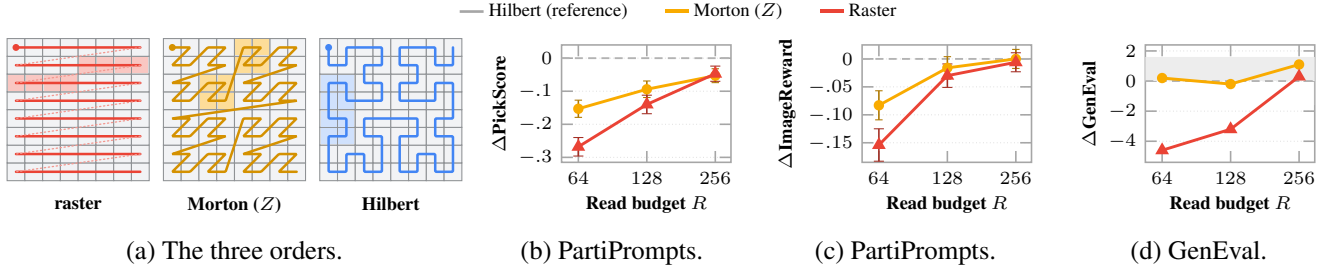
\begin{figure*}[t]
\centering
\pgfplotsset{ordax/.style={
    width=\linewidth, height=0.82\linewidth,
    xmin=54, xmax=300, xtick={64,128,256}, xticklabels={$64$,$128$,$256$},
    xlabel={\scriptsize\textbf{Read budget $\Rcells$}},
    ylabel style={font=\scriptsize}, xlabel style={font=\scriptsize},
    tick label style={font=\scriptsize}, xlabel shift=-2pt, ylabel shift=-2pt,
    ymajorgrids=true, grid style={line width=0.3pt, draw=gray!30, densely dotted},
    axis line style={line width=0.6pt, draw=gray!70},
    tick style={draw=gray!70, line width=0.6pt}, tick align=inside, minor tick num=0,
    scaled y ticks=false, clip=false,
}}
\centering
{\scriptsize\color{black}
  \textcolor{gray!70}{\rule[0.35ex]{9pt}{1.0pt}}\;Hilbert (reference)\qquad
  \textcolor{gyellow}{\rule[0.35ex]{9pt}{1.4pt}}\;Morton ($Z$)\qquad
  \textcolor{gred}{\rule[0.35ex]{9pt}{1.4pt}}\;Raster}\\[3pt]
\begin{minipage}[b]{0.34\textwidth}
\centering
\begin{tikzpicture}[font=\scriptsize, x=0.235cm, y=0.235cm,
  opath/.style={line width=0.95pt, rounded corners=0.5pt, line cap=round}]
\begin{scope}[shift={(0,0,0)}]
  \foreach \i in {0,...,7}{\foreach \j in {0,...,7}{
    \draw[draw=ggreyd!70, fill=ggrid!55, line width=0.4pt] (\i,\j) rectangle ++(1,1);}}
  \foreach \c in {(4,6),(5,6),(6,6),(7,6),(0,5),(1,5),(2,5),(3,5)}
    \fill[gred!30] \c ++(0.06,0.06) rectangle ++(0.88,0.88);
  \foreach \y in {0,...,7}
    \draw[gred, opath] (0.5,\y+0.5) -- (7.5,\y+0.5);
  \foreach \y in {1,...,7}
    \draw[gred!55, line width=0.5pt, densely dotted, line cap=round]
      (7.5,\y+0.5) -- (0.5,\y-0.5);
  \fill[gred] (0.5,7.5) circle (1.3pt);
  \node[text=black, font=\scriptsize\bfseries, anchor=north] at (4,-0.4) {raster};
\end{scope}
\begin{scope}[shift={(8.8,0,0)}]
  \foreach \i in {0,...,7}{\foreach \j in {0,...,7}{
    \draw[draw=ggreyd!70, fill=ggrid!55, line width=0.4pt] (\i,\j) rectangle ++(1,1);}}
  \foreach \c in {(2,5),(3,5),(2,4),(3,4),(4,7),(5,7),(4,6),(5,6)}
    \fill[gyellow!40] \c ++(0.06,0.06) rectangle ++(0.88,0.88);
  \draw[gyellow!85!black, opath] plot coordinates {
    (0.5,7.5)(1.5,7.5)(0.5,6.5)(1.5,6.5)(2.5,7.5)(3.5,7.5)(2.5,6.5)(3.5,6.5)
    (0.5,5.5)(1.5,5.5)(0.5,4.5)(1.5,4.5)(2.5,5.5)(3.5,5.5)(2.5,4.5)(3.5,4.5)
    (4.5,7.5)(5.5,7.5)(4.5,6.5)(5.5,6.5)(6.5,7.5)(7.5,7.5)(6.5,6.5)(7.5,6.5)
    (4.5,5.5)(5.5,5.5)(4.5,4.5)(5.5,4.5)(6.5,5.5)(7.5,5.5)(6.5,4.5)(7.5,4.5)
    (0.5,3.5)(1.5,3.5)(0.5,2.5)(1.5,2.5)(2.5,3.5)(3.5,3.5)(2.5,2.5)(3.5,2.5)
    (0.5,1.5)(1.5,1.5)(0.5,0.5)(1.5,0.5)(2.5,1.5)(3.5,1.5)(2.5,0.5)(3.5,0.5)
    (4.5,3.5)(5.5,3.5)(4.5,2.5)(5.5,2.5)(6.5,3.5)(7.5,3.5)(6.5,2.5)(7.5,2.5)
    (4.5,1.5)(5.5,1.5)(4.5,0.5)(5.5,0.5)(6.5,1.5)(7.5,1.5)(6.5,0.5)(7.5,0.5)};
  \fill[gyellow!85!black] (0.5,7.5) circle (1.3pt);
  \node[text=black, font=\scriptsize\bfseries, anchor=north] at (4,-0.4) {Morton ($Z$)};
\end{scope}
\begin{scope}[shift={(17.6,0,0)}]
  \foreach \i in {0,...,7}{\foreach \j in {0,...,7}{
    \draw[draw=ggreyd!70, fill=ggrid!55, line width=0.4pt] (\i,\j) rectangle ++(1,1);}}
  \foreach \c in {(1,4),(1,5),(0,5),(0,4),(0,3),(1,3),(1,2),(0,2)}
    \fill[gblue!25] \c ++(0.06,0.06) rectangle ++(0.88,0.88);
  \draw[gblue, opath] plot coordinates {
    (0.5,7.5)(0.5,6.5)(1.5,6.5)(1.5,7.5)(2.5,7.5)(3.5,7.5)(3.5,6.5)(2.5,6.5)
    (2.5,5.5)(3.5,5.5)(3.5,4.5)(2.5,4.5)(1.5,4.5)(1.5,5.5)(0.5,5.5)(0.5,4.5)
    (0.5,3.5)(1.5,3.5)(1.5,2.5)(0.5,2.5)(0.5,1.5)(0.5,0.5)(1.5,0.5)(1.5,1.5)
    (2.5,1.5)(2.5,0.5)(3.5,0.5)(3.5,1.5)(3.5,2.5)(2.5,2.5)(2.5,3.5)(3.5,3.5)
    (4.5,3.5)(5.5,3.5)(5.5,2.5)(4.5,2.5)(4.5,1.5)(4.5,0.5)(5.5,0.5)(5.5,1.5)
    (6.5,1.5)(6.5,0.5)(7.5,0.5)(7.5,1.5)(7.5,2.5)(6.5,2.5)(6.5,3.5)(7.5,3.5)
    (7.5,4.5)(7.5,5.5)(6.5,5.5)(6.5,4.5)(5.5,4.5)(4.5,4.5)(4.5,5.5)(5.5,5.5)
    (5.5,6.5)(4.5,6.5)(4.5,7.5)(5.5,7.5)(6.5,7.5)(6.5,6.5)(7.5,6.5)(7.5,7.5)};
  \fill[gblue] (0.5,7.5) circle (1.3pt);
  \node[text=black, font=\scriptsize\bfseries, anchor=north] at (4,-0.4) {Hilbert};
\end{scope}
\end{tikzpicture}
\par\smallskip
{(a) The three orders.}
\end{minipage}\hfill
\begin{minipage}[b]{0.215\textwidth}
\centering
\begin{tikzpicture}
  \begin{axis}[ordax, xmode=log, log basis x=2, ymin=-0.315, ymax=0.04,
    ytick={0,-0.1,-0.2,-0.3}, yticklabels={$0$,$-.1$,$-.2$,$-.3$},
    ylabel={\scriptsize\textbf{$\Delta$PickScore}}]
  \addplot[gray!60, densely dashed, line width=0.7pt] coordinates {(54,0)(300,0)};
  \addplot[gyellow, line width=1.1pt, mark=*, mark size=1.4pt, mark options={fill=gyellow},
    error bars/.cd, y dir=both, y explicit, error bar style={gyellow!70!black, line width=0.6pt}]
    coordinates {(64,-0.153) +- (0,0.026) (128,-0.094) +- (0,0.025) (256,-0.053) +- (0,0.021)};
  \addplot[gred, line width=1.1pt, mark=triangle*, mark size=1.9pt, mark options={fill=gred},
    error bars/.cd, y dir=both, y explicit, error bar style={gred!70!black, line width=0.6pt}]
    coordinates {(64,-0.268) +- (0,0.028) (128,-0.140) +- (0,0.028) (256,-0.047) +- (0,0.023)};
  \end{axis}
\end{tikzpicture}
\par\smallskip
{(b) PartiPrompts.}
\end{minipage}\hfill
\begin{minipage}[b]{0.215\textwidth}
\centering
\begin{tikzpicture}
  \begin{axis}[ordax, xmode=log, log basis x=2, ymin=-0.185, ymax=0.025,
    ytick={0,-0.05,-0.10,-0.15}, yticklabels={$0$,$-.05$,$-.10$,$-.15$},
    ylabel={\scriptsize\textbf{$\Delta$ImageReward}}]
  \addplot[gray!60, densely dashed, line width=0.7pt] coordinates {(54,0)(300,0)};
  \addplot[gyellow, line width=1.1pt, mark=*, mark size=1.4pt, mark options={fill=gyellow},
    error bars/.cd, y dir=both, y explicit, error bar style={gyellow!70!black, line width=0.6pt}]
    coordinates {(64,-0.083) +- (0,0.026) (128,-0.016) +- (0,0.020) (256,0.000) +- (0,0.017)};
  \addplot[gred, line width=1.1pt, mark=triangle*, mark size=1.9pt, mark options={fill=gred},
    error bars/.cd, y dir=both, y explicit, error bar style={gred!70!black, line width=0.6pt}]
    coordinates {(64,-0.154) +- (0,0.029) (128,-0.030) +- (0,0.021) (256,-0.006) +- (0,0.017)};
  \end{axis}
\end{tikzpicture}
\par\smallskip
{(c) PartiPrompts.}
\end{minipage}\hfill
\begin{minipage}[b]{0.215\textwidth}
\centering
\begin{tikzpicture}
  \begin{axis}[ordax, xmode=log, log basis x=2, ymin=-5.4, ymax=2.4,
    ytick={2,0,-2,-4}, yticklabels={$2$,$0$,$-2$,$-4$},
    ylabel={\scriptsize\textbf{$\Delta$GenEval}}]
  \addplot[draw=none, fill=gray!14, forget plot]
    coordinates {(54,-1.6) (54,1.6) (300,1.6) (300,-1.6)} \closedcycle;
  \addplot[gray!60, densely dashed, line width=0.7pt] coordinates {(54,0)(300,0)};
  \addplot[gyellow, line width=1.1pt, mark=*, mark size=1.4pt, mark options={fill=gyellow}]
    coordinates {(64,0.2)(128,-0.2)(256,1.1)};
  \addplot[gred, line width=1.1pt, mark=triangle*, mark size=1.9pt, mark options={fill=gred}]
    coordinates {(64,-4.6)(128,-3.2)(256,0.3)};
  \end{axis}
\end{tikzpicture}
\par\smallskip
{(d) GenEval.}
\end{minipage}
\caption{\textbf{The partition order matters at small budgets.}
\textbf{(a)}~The three orders on an $8{\times}8$ example grid, with the same run of eight
consecutive positions shaded in each --- the mean region at $\Rcells{=}128$. A raster run
is a row strip that wraps at the row end (dotted); a Morton run joins two dyadic blocks
by a diagonal jump; a Hilbert run is a compact $2{\times}4$ patch. Only Hilbert keeps
every run connected. \textbf{(b,c)}~Each order retrains the MiniT2I-B/16 recipe with nothing else changed
and is read against the matched Hilbert run; $\Delta$ is order minus Hilbert on
PartiPrompts. The ranking follows the locality of the order, and the deficit grows as the
budget shrinks. Error bars: paired-bootstrap $95\%$ CIs over the $1632$ prompts. \textbf{(d)}~GenEval against its own references
(gray band: $\pm1.6$ run-to-run spread). Raster falls far outside the band at small
budgets, while Morton stays at the reference: composition survives Morton's dyadic jumps
but not raster's row wraps.}
\label{fig:ordering}
\end{figure*}

\begin{figure*}[t]
\centering
\input{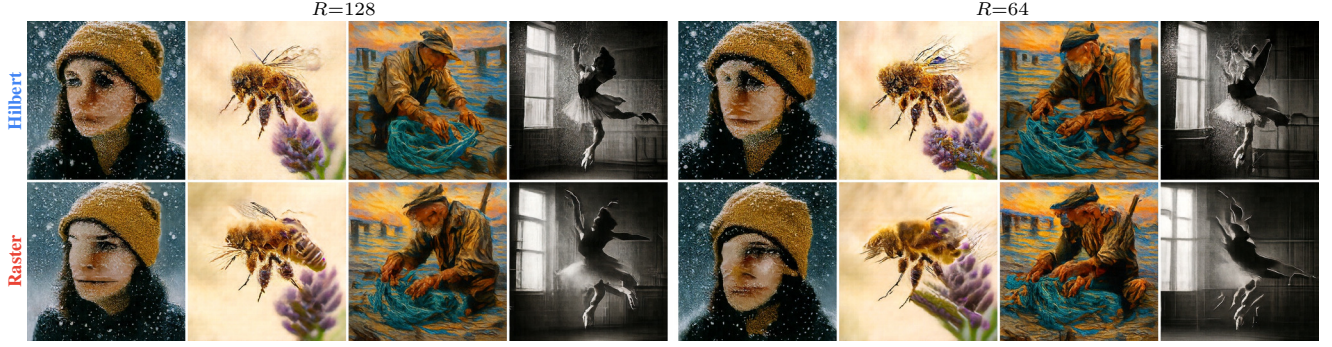}
\caption{\textbf{The ordering ablation, qualitatively}: We compare two orderings visually on MiniT2I-B/16 --- Hilbert and raster --- at extreme budgets $\Rcells{=}128$ (left block)
and $\Rcells{=}64$ (right block), same prompt and seed within a column. Consistent with
\cref{fig:ordering}b--d, the orderings are close at $\Rcells{=}128$; at $\Rcells{=}64$ Hilbert keeps the face, the wings and the mesh where raster smears them, and holds the
dancer together.}
\label{fig:orders-samples}
\end{figure*}

\begin{table*}[t]
\begin{minipage}[t]{0.485\textwidth}
\vspace{0pt}
\centering\footnotesize
\setlength{\tabcolsep}{3pt}
\begin{tabularx}{\linewidth}{@{}lXcccc@{}}
\toprule
 & core & $\Rcells$ & cost & GenEval\,$\uparrow$ & Pick\,$\uparrow$ \\
\midrule
\multirow{2}{*}{\emph{same budget}}
 & $[3,13]$ & $256$ & $1.00$ & $\mathbf{84.7}$ & $\mathbf{22.10}$ \\
 & $[1,15]$ & $256$ & $0.66$ & $77.4$ & $21.70$ \\
\addlinespace[2pt]
\multirow{2}{*}{\emph{same cost}}
 & $[3,13]$ & $256$ & $1.00$ & $84.7$ & $22.10$ \\
 & $[1,15]$ & $512$ & $\approx1$ & $85.4$ & $22.14$ \\
\bottomrule
\end{tabularx}
\caption{\textbf{The core length is a compute knob, not a quality knob} (B/16,
text-to-image). The core of the paper spans $11$ blocks ($[3,13]$); the alternative
spans $15$ ($[1,15]$) and so leaves two full-resolution blocks instead of six.
\emph{Cost} is the image-token cost of one forward pass relative to the first row. At
the same budget the longer core is worse on every metric --- but also a third cheaper,
so that comparison is not a fair one. At the budget that restores its cost ($\Rcells{=}512$, inside the iso-cost window
$\Rcells{\approx}461$--$572$) the two are indistinguishable: GenEval $+0.7$ lies inside
the $\pm1.7$ interval of the reference row, and PickScore $+0.037$ (paired-bootstrap CI
$[+0.013,+0.061]$) is $0.17\%$ of its scale. ImageReward is omitted for space and moves
by $+0.002$ (CI $[-0.015,+0.019]$), an interval spanning zero. The length buys compute,
not quality.}
\label{tab:span}
\end{minipage}\hfill
\begin{minipage}[t]{0.485\textwidth}
\vspace{0pt}
\centering\footnotesize
\setlength{\tabcolsep}{3pt}
\begin{tabularx}{\linewidth}{@{}Xccc@{}}
\toprule
 & \emph{contiguous} & \emph{content-sized} & \emph{summarizes} \\
 & spread\,$\downarrow$ & size\,--\,detail\,$\uparrow$ & retained\,$\uparrow$ \\
\midrule
\method{} (ours)   & $\mathbf{0.49}$ & $\mathbf{0.50}$ & $0.73$ \\
fixed boundaries   & $0.71$ & $0.00$ & $0.67$ \\
Feat Sim           & $5.16$ & $0.03$ & $\mathbf{0.74}$ \\
Latent Array       & $11.35$ & $0.20$ & $-0.08$ \\
Token Skip         & \multicolumn{2}{c}{\emph{no groups}} & $0.26$ \\
\midrule
\emph{whole image} & $12.24$ & --- & --- \\
\bottomrule
\end{tabularx}
\caption{\textbf{The three measurements behind \cref{fig:mechanism}}, at $\Rcells{=}256$
over $64$ prompts. Spread: mean distance from a patch to its group centre, in cells of the $32{\times}32$ grid ($12.24$: a group covering the whole image).
Size--detail: correlation between a group's fineness and the detail it covers (fixed
boundaries give $0$ by construction). Retained: share of the core representation that
survives the reduction, without any learned \Write{}. The latent array's $-0.08$: its pooled tokens reconstruct the block worse than the
global mean. The similarity merge retains the most and still loses on quality --- it
gives up position, not information.}
\label{tab:properties}
\end{minipage}
\end{table*}

\begin{figure*}[t]
\centering
\input{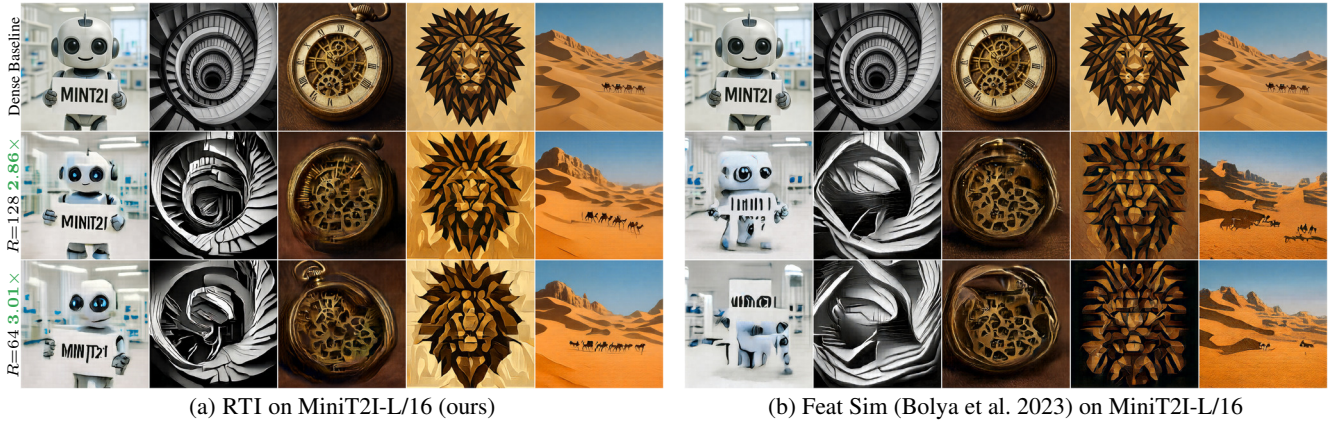}
\caption{\textbf{Failure cases at the two smallest budgets, MiniT2I-L/16}, in the
layout of \cref{fig:gallery}: the frozen backbone (top row of each panel) against
\textbf{(a)}~the elastic \method{} checkpoint and \textbf{(b)}~the Feat Sim twin, read
at $\Rcells{=}128$ and $\Rcells{=}64$. Same prompt and seed within a column, identical token counts and throughput per budget
row.}
\label{fig:failcase}
\end{figure*}

\begin{figure*}[t]
\centering
\input{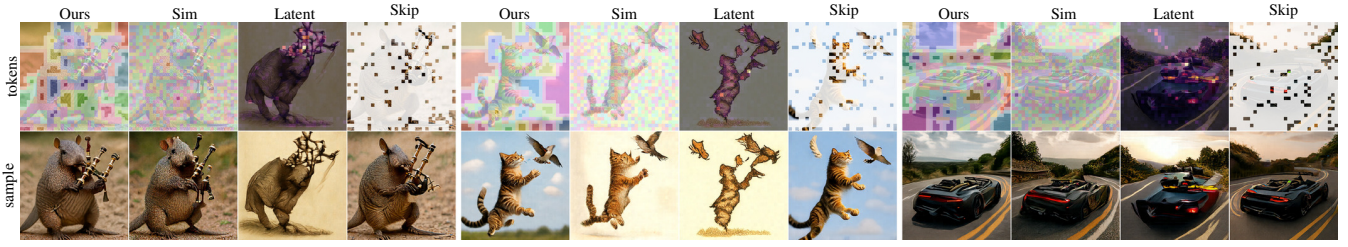}
\caption{\textbf{The three reductions at the same budget $\Rcells{=}256$}, three prompts
in blocks of four (\emph{Ours}, Feat \emph{Sim}, \emph{Latent} array, Token
\emph{Skip}). Top: what each method does to the $\Ntok{=}1024$ patch tokens; bottom: the sample from the same run, as in \cref{fig:mechanism}; each method's own
trained checkpoint, same prompt and seed.}
\label{fig:method_samples}
\end{figure*}

\section{Technical Details}
\label{sec:supp-impl}
\label{sec:supp-baselines}

Every model in the paper adapts a released checkpoint that remains frozen. Training
updates only the \Read{}/\Write{} interface and rank-$32$ LoRA adapters on the attention
and MLP projections of the trunk. The budget $\Rcells$ is drawn anew at every training
step, so a single checkpoint serves the whole budget range. The text stream always runs
at full length, and the partition is rebuilt from the current state at every denoising
step, so the regions follow the image as it forms. \cref{tab:settings} lists all
settings.

\paragraph{Baseline implementations.} All baselines are our implementations inside the
same retrofit: the same frozen backbone, LoRA rank and placement, recipe, data, and
budget protocol where the method has a budget. Only the token-reduction rule differs.
\emph{Feat Sim} replaces only the partition. Following ToMe~\citep{bolya2023tome}, it
picks $\Rcells$ evenly spaced anchor tokens and assigns every patch to its nearest anchor
by cosine similarity; \Read{} and \Write{} then act on these similarity groups exactly as
on our regions. Its budget is also drawn per training step, so this checkpoint is elastic
as well and runs at the same token count and throughput as ours at every budget.
\emph{Latent Array} removes the partition. Here, $\Rcells$ learned latents read the patch
sequence by cross-attention before the core and write it back afterwards, as in
Perceiver-IO and RIN~\citep{jaegle2022perceiverio,jabri2023rin}; the latents carry no
rotary position. \emph{Token Skip} follows Mixture-of-Depths~\citep{raposo2024mod}. Its
middle blocks alternate between dense and sparse; a learned router admits the top
$12.5\%$ of image tokens to each sparse block, and the remaining tokens bypass it on the
residual stream. It has no budget parameter, so we report it at its capacity $C{=}0.125$,
which corresponds to $576$ image tokens per middle block on average (footnoted in
\cref{tab:main}). Its dense/sparse alternation covers blocks $[2,13]$ rather than our
core $[3,13]$.

\paragraph{Compute accounting.} \cref{fig:flops} compares the analytic cost of one
forward pass with its measured wall-clock time on B/16, both relative to the dense
forward. The FLOP count covers the trunk's linear layers and the joint attention over
image and text tokens, with the text stream at full length in every block; the partition
and \Read{}/\Write{} are $\mathcal{O}(\Ntok d)$ and contribute less than $2\%$. The measured time lies below the FLOP curve at every budget, which is why the paper
reports measured throughput rather than FLOPs.

\section{Additional Results}
\label{sec:supp-exp}

\subsection{FID on MJHQ-30K}
\label{sec:supp-mjhq}

GenEval and the preference models score each image individually, whereas
FID~\citep{heusel2017fid} measures whether the generated distribution stays close to a
reference. We
compute it on MJHQ-30K~\citep{li2024playground} for L/16, with $30{,}000$ prompts and one
image per prompt, under the protocol of \cref{sec:exp-setup}, using
clean-fid~\citep{parmar2022cleanfid} against the official statistics (\cref{fig:fid}a).
Feat Sim is read at the same budgets, so at each speedup the two differ only in the
grouping rule.

At matched budget and throughput, \method{} is better than Feat Sim by $8.6$ FID at
$\Rcells{=}512$ and by $13.3$ at $\Rcells{=}256$; the gap widens as the budget shrinks,
in line with the position argument of \cref{sec:method}. \method{} at $\Rcells{=}256$ also reaches a lower FID
than Feat Sim at $\Rcells{=}512$, with half the tokens and a higher speedup. FID is, however, the one metric on which \method{} does not match the dense model at
half the patch tokens: $+2.4$ at $\Rcells{=}512$, where GenEval is within noise. Compression therefore costs
distributional fidelity before it costs compositional accuracy, consistent with the
class-conditional results below. One caveat applies: guidance $5.0$ follows the
protocol of the paper but is high for FID; it inflates every method equally, so the comparisons hold, but the absolute values should not be compared with guidance-tuned
numbers.

\subsection{Class-Conditional Generation}
\label{sec:supp-c2i}

We next test the interface on a different task and backbone family. The recipe carries
over --- frozen backbone, LoRA, the same Hilbert partition, random training budgets ---
to class-conditional ImageNet-$256$ with the pretrained pixel-space JiT-B/16 and
JiT-L/16 backbones~\citep{liBackBasicsLet2026}; only the core placement follows the new
backbone, as discussed below. We evaluate under the protocol of that work:
$50$K samples, guidance $3.0$ on $[0.1,1.0]$, $50$ Heun steps. The grid has $16{\times}16{=}256$ tokens, so $\Rcells{\in}\{64,130,193\}$ is $25/50/75\%$
of them and $62/75/88\%$ of the dense token compute. The checkpoint trains on $\Rcells{\in}\{32,64,128,192\}$, so $130$ and $193$ also probe
budgets it never saw --- at no cost: the trained $\Rcells{=}128$ gives $6.34$ FID against
$6.23$ at $130$, so quality follows the budget smoothly rather than peaking at the
trained values.

\method{} is better than Feat Sim at every budget, on both backbones, in FID and
Inception Score, and the margin is largest at intermediate budgets: $2.6\times$ lower FID
at $\Rcells{=}130$ on B/16 ($6.23$ vs $15.99$), the regime where the budget binds but the
image is still recoverable. Against Token Skip it wins while spending less: at
$\Rcells{=}64$ it uses $62\%$ of the dense token compute against Token Skip's $71\%$ and
reaches $16.69$ against $19.11$ FID, and at $\Rcells{=}130$ it beats the depth-augmented
variant ($6.23$ against $7.02$) at three quarters of that variant's token compute. Summarizing a
region is therefore better than dropping or skipping its tokens on a second task, a
second backbone family, and a distributional metric.

On the larger backbone the compressed model reaches the dense one, at three quarters of
the patch tokens rather than half. At $\Rcells{=}193$,
which spends $88\%$ of the dense token compute, \method{} scores $2.42$ FID against the published $2.36$ of the dense
JiT-L/16~\citep{liBackBasicsLet2026}, and it is above the dense model on Inception Score at that
budget and at $\Rcells{=}130$ ($313.5$ and $303.8$ against $298.5$). A higher Inception
Score is not by itself evidence of better images, since the metric rewards confident,
prototypical class predictions; the FID column carries the claim.

\paragraph{Core placement.} The c2i core $[4,9]$ is a structural choice. JiT-B/16
prepends its $32$ in-context class tokens at block $4$, so the core starts where the
class conditioning enters, and the patch-only encoder blocks stay at full resolution.
\cref{fig:poolability-jit} checks whether the redundancy profile constrains this choice,
by running the probe of \cref{sec:supp-poolability} on five frozen JiT backbones. The
profile does not constrain it, and two observations follow. First, the redundancy the interface exploits is present
in every backbone: $25\%$ of the tokens retain about $70\%$ of the representation.
Second, the hourglass shape grows with resolution and model size rather than with the
task. At $256^2$ it is present but shallow: the rise from the first block to the peak is
about $0.03$, against $0.09$--$0.13$ at $512^2$, where one patch covers four times as
many pixels. A profile that shallow leaves the core placement effectively free ---
longer candidate cores have the same mean EV --- so $[4,9]$ is a conservative choice
aligned with the model structure rather than one the profile forces. A c2i run with a longer core remains an open and cheap experiment.

\subsection{Fewer Sampling Steps}
\label{sec:supp-steps}

Fewer sampling steps and a smaller region budget both save wall-clock time, so we compare
them at equal cost on L/16. One trajectory costs the measured time of a forward pass
($288$\,ms dense, $112$\,ms at $\Rcells{=}256$) times the number of steps;
\cref{fig:steps-supp} sweeps all four metrics and pairs the two options wherever their
costs match to within $4\%$. All other settings follow \cref{sec:exp-setup}.

There are two regimes. Below about three seconds per image, \method{} is clearly ahead
($82.7$ vs $65.9$ GenEval at $1.8$\,s): truncating the sampler below roughly ten steps
destroys the image, while compressing tokens degrades it gradually. Read at equal
quality, \method{} is $2.1\times$ cheaper. Above about five seconds, additional steps are
the better investment and the dense model ends ahead ($87.6$ vs $86.3$ GenEval at
$11.2$\,s). Region compression therefore extends the usable cost range of the backbone without
dominating step reduction everywhere, and the $100$-step setting of \cref{tab:main}
matches dense quality without being the compute optimum. Two qualifications: six-step Euler is a weak sampler, so a higher-order solver would
narrow the low-end gap; and the winning configurations are themselves short
trajectories, so a region interface on a few-step distilled model is a natural
combination.

\subsection{Ablation on Partition Order}
\label{sec:supp-ordering}

\Cref{sec:supp-property} compares token orders on the \emph{frozen} model; here we train
them in. \cref{fig:ordering}a shows the three candidates. The partition always cuts the
chosen order into runs that are contiguous along the order; the orders differ in whether
this contiguity survives in the image. A Hilbert run is connected by construction --- the walk only steps between grid
neighbours --- though not necessarily square. A raster run tears at row ends, and a Morton run at dyadic-block
boundaries. The swap therefore isolates the contiguity property of \cref{sec:intro}.
Starting from the MiniT2I-B/16 recipe, we replace the Hilbert order by the raster or
Morton ($Z$) order, change nothing else ($40$K steps), and read each order against the matched Hilbert
reference at three budgets (\cref{fig:ordering}b--d).

The effect is budget-dependent, and the mechanism is run length. At $\Rcells{=}256$ a
mean region holds four patches; a run of four stays compact under any of the three
orders, and all differences lie within noise. As the budget shrinks, a region grows to eight and then sixteen patches, the order
decides more of what a run covers, and the ranking becomes monotone in its locality ---
as \cref{sec:method} predicts and the frozen-model ranking of \cref{sec:supp-property}
already showed. At $\Rcells{=}64$ the effect is clearly significant: paired
bootstraps over the $1632$ prompts place raster at $-0.268$ PickScore (CI
$[-0.296,-0.239]$) and $-0.154$ ImageReward (CI $[-0.182,-0.125]$) below Hilbert, with
Morton in between ($-0.153$ / $-0.083$); every interval excludes zero. GenEval
(\cref{fig:ordering}d) separates the two failure modes: raster loses $3.2$ and $4.6$
points at $\Rcells{=}128$ and $64$, far outside the $\pm1.6$ spread, while Morton stays
at the reference at every budget. Tearing runs at row ends damages composition; Morton's
dyadic jumps preserve composition and cost only perceptual quality.
\cref{fig:orders-samples} shows the two extremes; Morton, visually close to Hilbert, is
omitted there. In short, the order does not matter at rich budgets and matters at lean
ones, and Hilbert --- the only order whose runs stay connected --- is the natural
choice.

\subsection{Lengthening the Core}
\label{sec:supp-span}

This study is on the text-to-image backbone MiniT2I-B/16, not on the
class-conditional JiT of \cref{sec:supp-c2i}. We lengthen its core from $[3,13]$ to
$[1,15]$ and change nothing else (\cref{tab:span}). At matched budget the longer core is worse on every metric. It is,
however, also cheaper: $0.66\times$ the image-token cost at $\Rcells{=}256$
($0.44\times$ on attention), which corresponds to an iso-cost budget of
$\Rcells{\approx}461$--$572$. Paired at matched compute, the two cores are indistinguishable on every metric
(\cref{tab:span}). Neither length is better: the core length buys compute, and a larger budget spends it
to the same effect. $[3,13]$ is an operating point on that frontier, not an optimum, consistent with the depth
profile of \cref{fig:poolability}c. Two caveats: the iso-cost window counts image tokens only (the
full-length text stream pushes the true iso-throughput budget below $\Rcells{=}461$),
and $[1,15]$ was not timed --- with only two full-resolution blocks, its speedup ceiling
lies above the $2.5\times$ saturation of \cref{sec:exp-main}.

\section{Token Redundancy and Region Analysis}
\label{sec:supp-probes}

This section gives the protocols behind \cref{fig:poolability} (how redundant the tokens
of a pretrained model are, and where), \cref{fig:mechanism} and the property costs of
\cref{sec:exp-main} (how the groupings compare on the same features), and
\cref{fig:routing} (where the trained model places its partitions).

\paragraph{Shared setup.} The first two analyses run the released MiniT2I-B/16 without
adapters or interface, generating at $512^2$ in bf16 with $30$ Euler steps at guidance
$5.0$. The third runs a trained elastic checkpoint, since it describes partitions the
model builds for itself. All three use held-out prompts:
synthetic templates over ten domains, filtered for near-duplicates against GenEval and
PartiPrompts. Each averages over prompts and over steps spread across the trajectory ($128$ prompts at
$8$ steps, $64$ at $4$, $48$ at $10$). Hidden states are captured with hooks: every block for the first analysis, the input to
block $s_0{=}3$ for the other two --- what the partition is built from at inference. Only image tokens are measured; the $\Ntok{=}1024$ of them form a $32{\times}32$ grid, and
all distances below are in grid cells. Our groups are runs of the Hilbert order, Feat Sim's of the raster order, each paired
with the positions of its own order.

\subsection{Measuring the Redundancy}
\label{sec:supp-poolability}

Given a grouping into $\Rcells$ groups, we replace each token by its group mean and
measure the retained variance:
\begin{equation}
\mathrm{EV} = 1 - \frac{\sum_j \|\mathbf{h}_j - \bar{\mathbf{h}}_{c(j)}\|^2}
                       {\sum_j \|\mathbf{h}_j - \bar{\mathbf{h}}\|^2},
\label{eq:ev}
\end{equation}
with $\bar{\mathbf{h}}_{c(j)}$ the group mean and $\bar{\mathbf{h}}$ the global mean.
$\mathrm{EV}{=}1$: the pooled tokens carry the representation exactly; $0$: no more
than a constant. We compare three groupings, all at $\Rcells{=}256$
of $1024$: \emph{adaptive} (ours; the Hilbert order cut at the $\Rcells{-}1$ largest
feature gaps), \emph{fixed} (evenly spaced cuts on the same
order; at $\Rcells{=}256$ exactly the $2{\times}2$ cells of a coarsened grid), and \emph{skip} (not a grouping: keep the $\Rcells$ highest-scoring tokens, replace the
rest by the mean).

In \cref{fig:poolability}, panel~(a) sorts patches by the gradient energy of the
finished image and plots cumulative detail against the share of patches; a curve above
the diagonal means detail sits in few patches. The images are the $128$ samples above, so the measurement describes what the model
itself draws.
Panel~(b) shows $\mathrm{EV}$ at a middle block over sampling steps; panel~(c) shows
$\mathrm{EV}$ per block averaged over steps, with the full-resolution ends shaded.
$\mathrm{EV}$ measures linear headroom on one block's activations, not output quality;
hence the method is then trained and evaluated end to end. The
$26\%$ that pooling does not recover at $\Rcells{=}256$ is why \method{} keeps a
full-resolution coda and a learned \Write{}.

\subsection{Comparing the Groupings}
\label{sec:supp-property}

Every method forms its groups on the same frozen features: ours cuts the Hilbert order
at the largest gaps, \emph{fixed} spaces the cuts evenly, \emph{Feat Sim} assigns each
token to the nearest of $\Rcells$ evenly spaced anchors by cosine similarity, \emph{Latent Array} has no hard groups, so each latent's softmax read is treated as a
soft group with weights $w_{rj}$ (its read queries are learned, hence taken from its
trained checkpoint); \emph{Token Skip} forms no groups, so only the third
measurement applies to it.

\emph{Contiguity}: with weights $w_{rj}$ summing to one, a group's centre is
$\mathbf{p}_r = \sum_j w_{rj}\mathbf{p}_j$ and its spread
\begin{equation}
\mathrm{spread}_r = \textstyle\sum_j w_{rj}\, \|\mathbf{p}_j - \mathbf{p}_r\|_2 ,
\label{eq:spread}
\end{equation}
averaged over groups. A single patch scores $0$; a uniform draw over the grid $12.2$;
the same number answers anchoring, since a group in one place has a well-defined
position. \emph{Region size follows content}: a patch's detail is the
gradient energy of $\hat x$ pooled to the patch grid, and fineness is $-\log_2$ of the
group size (for a soft group, the size is its participation ratio $1/\sum_j w_{rj}^2$).
The column reports the correlation between fineness and detail across groups, read
mid-trajectory (near zero for every method at the noisiest steps); a fixed grid scores
exactly $0$. \emph{Representation retained}: \cref{eq:ev} on each method's own grouping, averaged per
block across the core $[3,13]$; the latent array pools with its attention weights and
scatters with the column-normalised transpose. No learned \Write{} takes part, ours
included, so the column measures the grouping rule alone. It is a lower bound, equally pessimistic for every method, and the protocol behind
\cref{fig:poolability}c; the two agree where they overlap ($0.73$ vs $0.74$ for ours,
$0.67$ both times for fixed).

\paragraph{Which order?} Contiguity is a property of the order a run is cut from, so we repeat the measurement
on the Hilbert, raster and Morton ($Z$) orders; the frozen base saw none of them, so
none is favoured. Hilbert retains the most of the core representation ($0.735$ vs $0.720$ for Morton and
$0.697$ for raster) and forms the most compact groups ($1.49$ grid cells vs $1.73$ and
$3.85$): the order whose runs stay compact describes the block best. Morton is the informative comparison: every dyadic cell is one of its runs too
(\cref{sec:prelim}), so the two differ only in its diagonal jumps against Hilbert's
neighbour steps --- worth $0.015$. For scale, cutting by content rather than uniformly is worth $0.059$, so the full spread
across orders ($0.038$, Hilbert over raster) is about two thirds of that. (Spread here is per patch, not per group as in \cref{tab:properties}.)

\subsection{Following the Trained Partition}
\label{sec:supp-routing}

Where do the fine regions of the trained partition go? Detail is again the gradient
energy of $\hat x$. A region is \emph{fine} at one patch, \emph{coarse} at sixteen or more. As the image
forms, fine regions hold about $6\times$ the mean detail of coarse ones ($1$ would be
chance), and fineness correlates with detail at $+0.58$ ($\Rcells{=}256$, mid-sampling).
Unlike the size--detail column of \cref{tab:properties}, computed on frozen features,
this reads the partition the trained model builds. Tiling variety is the patch-weighted entropy, in bits, of the region-size distribution.
These describe behaviour; the controlled comparison is the fixed-boundary swap of
\cref{sec:exp-ablations}.

\section{Additional Visual Results}
\label{sec:supp-visuals}

\cref{fig:failcase} pushes \method{} and Feat Sim to the two smallest trained budgets.
Both degrade here, but in different ways. At $\Rcells{=}128$ \method{} keeps the
readable elements of the scene: the sign still reads, and the facets of the geometric
lion stay separate. Feat Sim garbles the lettering, breaks the lion into loose stripes,
and turns the spiral staircase into smooth folds. At $\Rcells{=}64$ the sign stays partly readable under \method{} and is lost under Feat
Sim. Not every failure separates the two: both lose the watch dial
already at $\Rcells{=}128$, and both keep the smooth dunes of the last column even at
$\Rcells{=}64$.

\cref{fig:method_samples} extends \cref{fig:mechanism} to three prompts. The merge flattens the surroundings of the subject; the latent array carries no
position, and its subjects fall apart; token skipping keeps the salient object and thins the scene around it. The region
tokens keep both the object and its surroundings at the same budget.

Finally, \cref{fig:gallery-supp1,fig:gallery-supp2,fig:gallery-supp3,fig:gallery-supp4},
collected at the end of the supplement, extend the budget sweep of \cref{fig:gallery} to
twenty further prompts on MiniT2I-L/16, each read at $\Rcells{=}512$, $256$ and $128$
under a shared dense reference: two sheets of photographic and painterly scenes, one of
bright photographic and one of saturated scenes.

\begin{figure*}[p]
\centering
\input{figures/fig_gallery_supp}
\suppgallery{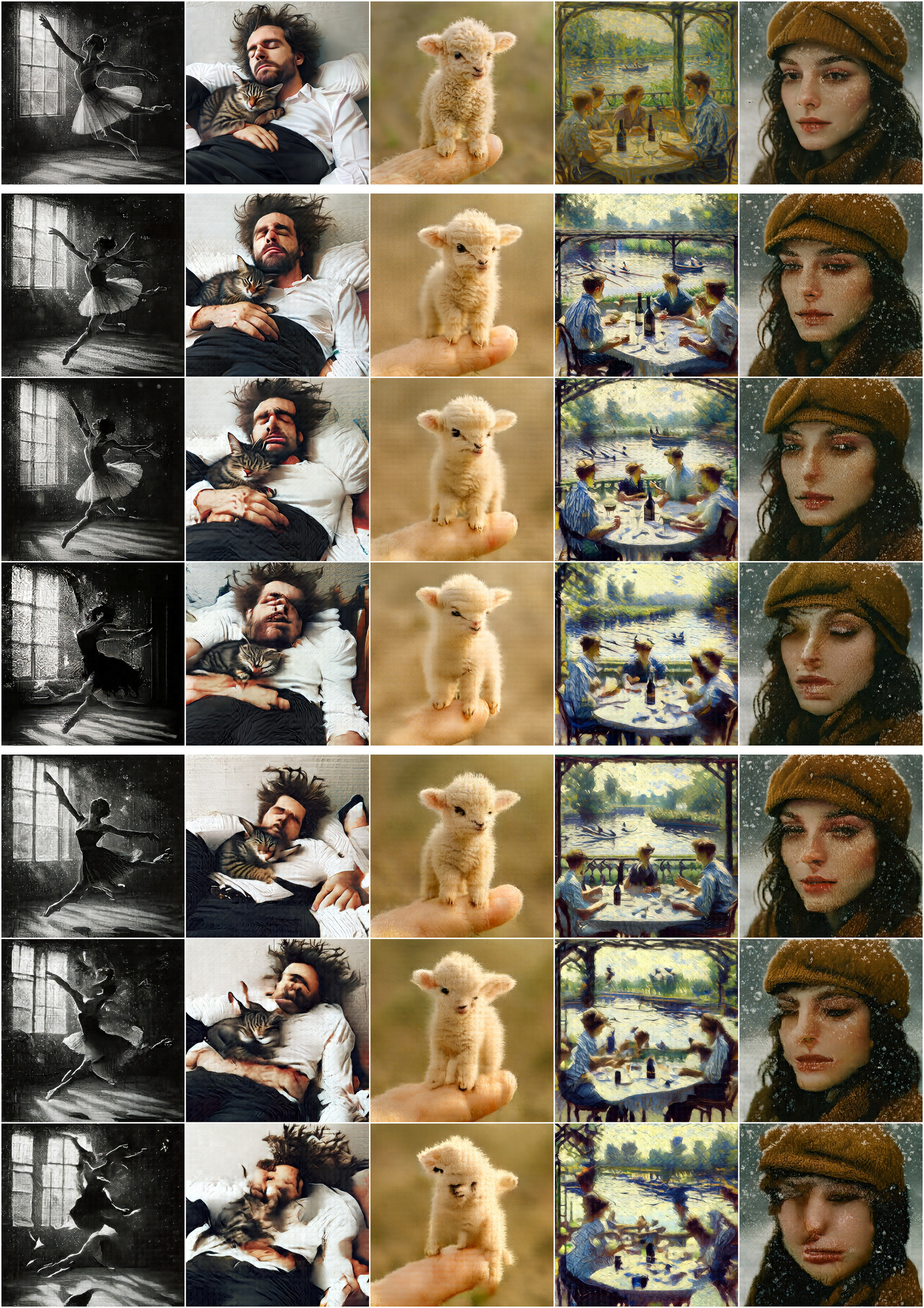}
\caption{\textbf{Budget sweep on MiniT2I-L/16} (sheet 1 of 4, protocol of
\cref{fig:gallery}). Top row: the frozen backbone. Below it, one elastic \method{}
checkpoint and the Feat Sim twin, each read at $\Rcells{=}512$ ($2.11\times$),
$\Rcells{=}256$ ($2.59\times$) and $\Rcells{=}128$ ($2.86\times$). Same prompt and seed within a
column; identical token counts and throughput for the two rows of a budget. Both methods
keep layout and palette throughout. The differences concentrate in faces and figures:
already at $\Rcells{=}512$ the merge alters the dancer's pose and the sleeping man's
face, and by $\Rcells{=}128$ it has dissolved the terrace into loose brushwork and
coarsened the snow portrait, where \method{} still holds them. On pure texture (the
lamb's wool) the two remain close at every budget.}
\label{fig:gallery-supp1}
\end{figure*}
\begin{figure*}[p]
\centering
\input{figures/fig_gallery_supp}
\suppgallery{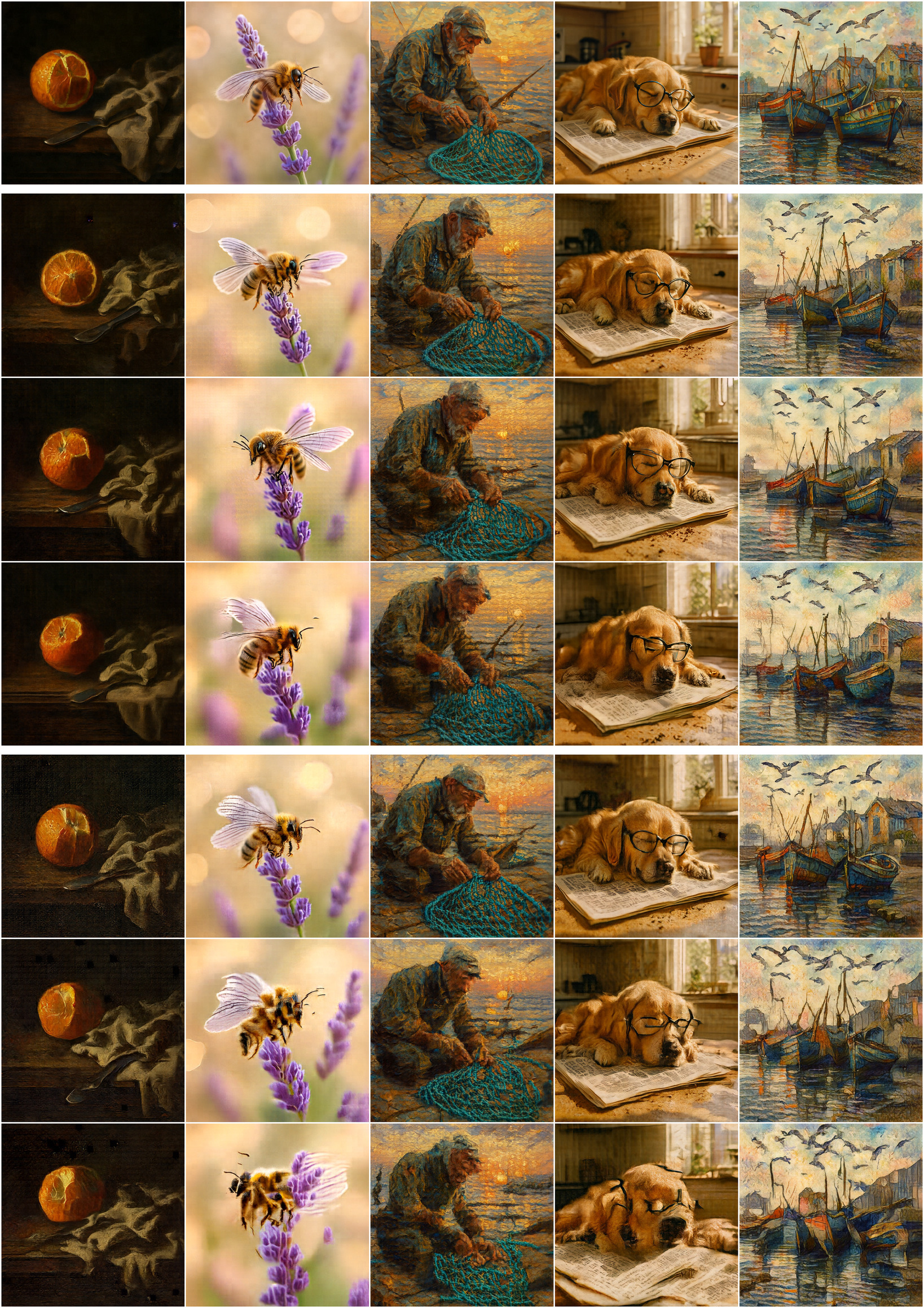}
\caption{\textbf{Budget sweep on MiniT2I-L/16} (sheet 2, same protocol as
\cref{fig:gallery-supp1}). At $\Rcells{=}256$ the merge un-peels and moves the orange,
doubles the bee's wing, turns the net's mesh into rope, and bends the dog's glasses and
the harbour's masts; \method{} keeps them. At $\Rcells{=}128$ both methods lose fine texture, but the merge also loses the objects
themselves, while \method{} keeps them
recognisable. The merge additionally drifts in composition, where \method{} stays close
to the dense row.}
\label{fig:gallery-supp2}
\end{figure*}
\begin{figure*}[p]
\centering
\input{figures/fig_gallery_supp}
\suppgallery{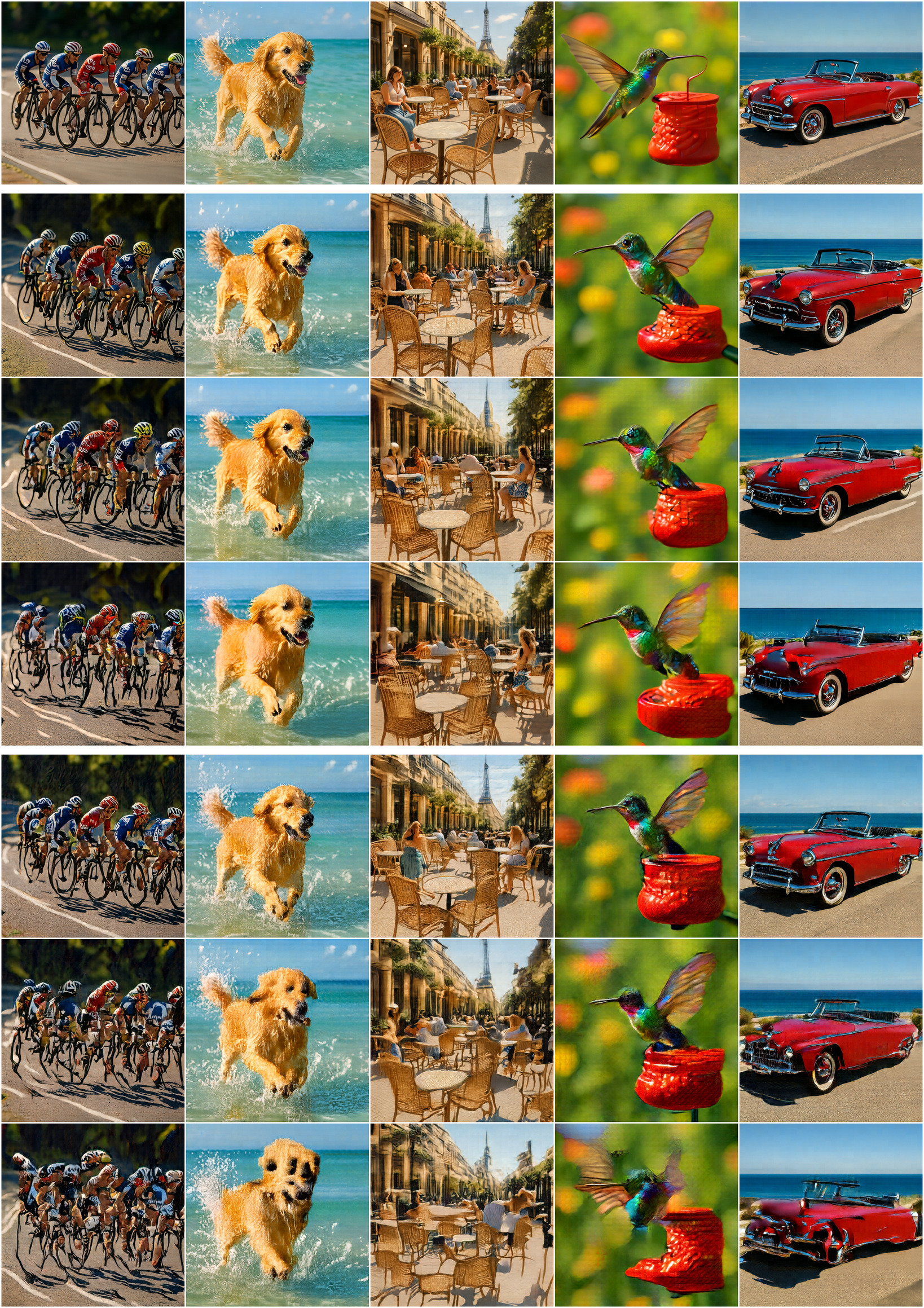}
\caption{\textbf{Budget sweep on MiniT2I-L/16} (sheet 3, same protocol as
\cref{fig:gallery-supp1}), on bright photographic scenes. The failures under the merge
are anatomical and mechanical rather than merely textural: at $\Rcells{=}256$ it fuses
the riders and their wheels in the peloton, deforms the retriever's head, multiplies and warps the rattan chairs, turns the hummingbird's feeder into a
shapeless object, and distorts the car's grille and wheels. At $\Rcells{=}128$ the retriever has
several faces and the feeder has lost its identity, while \method{} keeps one dog, one
feeder and one car at the same throughput.}
\label{fig:gallery-supp3}
\end{figure*}
\begin{figure*}[p]
\centering
\input{figures/fig_gallery_supp}
\suppgallery{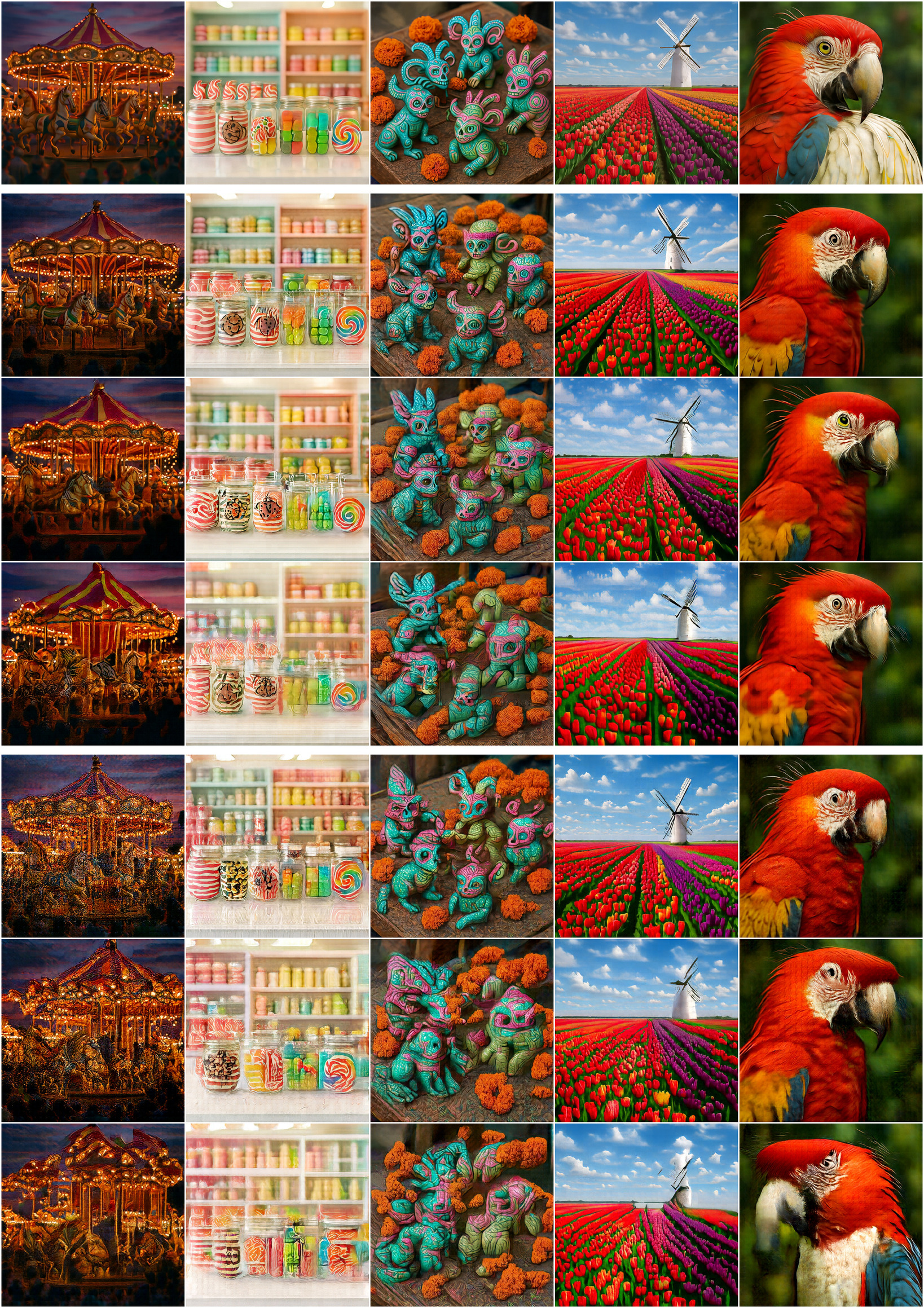}
\caption{\textbf{Budget sweep on MiniT2I-L/16} (sheet 4, same protocol as
\cref{fig:gallery-supp1}), saturated scenes. The merge fails earliest where colour sits
in small repeated elements: at $\Rcells{=}512$ it already speckles the carousel canopy
and scrambles the sweets in the jars; at $\Rcells{=}256$ it deforms the carousel horses,
dissolves the lollipop spirals, smears the alebrijes' painted patterns, and merges the
tulip rows. In the close-up of the macaw it smears the eye at $\Rcells{=}256$ and
deforms the whole head at $\Rcells{=}128$, where \method{} keeps the iris and the fine
lines of the bare facial skin. Colour itself survives in both; what is lost is the structure that carries it.}
\label{fig:gallery-supp4}
\end{figure*}

  \bibliography{main_lib,refs_extra}
\else
  \clearpage
  \appendix
  
\fi

\end{document}